\documentclass{article}

\usepackage{microtype}
\usepackage{graphicx}
\usepackage{subfigure}
\usepackage{booktabs} 

\usepackage{hyperref}

\usepackage[accepted]{icml2025}
\makeatletter
\renewcommand{\Notice@String}{} 
\makeatother

\usepackage{amsmath}
\usepackage{amssymb}
\usepackage{mathtools}
\usepackage{amsthm}

\usepackage[capitalize,noabbrev]{cleveref}

\theoremstyle{plain}

\theoremstyle{definition}

\theoremstyle{remark}

\usepackage[textsize=tiny]{todonotes}

\icmltitlerunning{The Impact of Machine Learning Uncertainty on the Robustness of Counterfactual Explanations}

\begin{document}

\twocolumn[
\icmltitle{The Impact of Machine Learning Uncertainty \\
on the Robustness of Counterfactual Explanations}



\icmlsetsymbol{equal}{*}

\begin{icmlauthorlist}

\icmlauthor{Leonidas Christodoulou}{yyy}
\icmlauthor{Chang Sun}{comp}

\end{icmlauthorlist}

\icmlaffiliation{yyy}{CaSToRC, The Cyprus Institute, Nicosia, Cyprus}
\icmlaffiliation{comp}{University of Bologna, Bologna, Italy}

\icmlcorrespondingauthor{Leonidas Christodoulou}{l.christodoulou@cyi.ac.cy}

\icmlkeywords{Machine Learning, ICML}

\vskip 0.3in
]



\printAffiliationsAndNotice{} 

\begin{abstract}
Counterfactual explanations are widely used to interpret machine learning predictions by identifying minimal changes to input features that would alter a model’s decision. However, most existing counterfactual methods have not been tested when model and data uncertainty change, resulting in explanations that may be unstable or invalid under real-world variability. In this work, we investigate the robustness of common combinations of machine learning models and counterfactual generation algorithms in the presence of both aleatoric and epistemic uncertainty.
Through experiments on synthetic and real-world tabular datasets, we show that counterfactual explanations are highly sensitive to model uncertainty. In particular, we find that even small reductions in model accuracy — caused by increased noise or limited data — can lead to large variations in the generated counterfactuals on average and on individual instances. These findings underscore the need for uncertainty-aware explanation methods in domains such as finance and the social sciences.
\end{abstract}

\section{Introduction}
\label{sec:intro}

Explainable AI (XAI) has become an essential component of research, development, and, increasingly, in the deployment of machine learning (ML) systems \cite{lipton2018mythos, burkart2021survey, molnar2022, Hassija2024, AzureCF}. The motivation for this is twofold. First, it is crucial for data science practitioners and end-users to understand how ML models operate, especially given their growing societal impact. Second, policy regulations such as the recently introduced EU AI act \cite{EUAIAct2024}, defines typical ML applications like creditworthiness as ``High-Risk AI Systems'', which consequently provide individuals with the right to explanation. This latter requirement, in practice, emphasizes the need for local explainability in ML. Over time, the field has converged on several standard methods for local explanations \cite{lime, lundberg2017unified, Wachter:2017aa}, among which counterfactual explanations have emerged as a particularly influential approach for interpreting both black-box and glass-box models \cite{Verma:2020aa, Kommiya_Mothilal_2021, Guidotti2022}. 

The appeal of counterfactual explanations (CE) lies in their alignment with the intuitive, "what-if" reasoning that humans commonly use to understand events in both science and everyday life. Rooted in long-standing traditions in philosophy and artificial intelligence \cite{Pearl_2009,morgan2014counterfactuals,scholkopf2021toward}, CE have more recently gained traction as a practical tool for explaining machine learning (ML) outcomes—often without a strict commitment to causal interpretation. This pragmatic shift accelerated following the influential work of Wachter et al. (2017) \cite{Wachter:2017aa}, who proposed a method for generating CE by jointly optimizing for minimal input changes (measured using the $\ell_1$ norm) and the desired counterfactual outcome. Their approach helped formalize the notion of proximity as a key criterion for CE quality, with the $\ell_1$ norm offering a clear, quantitative metric for evaluating how plausible or minimally invasive a CE is.

There is a plethora of other CE packages in the literature \cite{Guidotti2022}. For nearly all of them the main challenges is the time to run the code and how to handle categorical variables in tabular datasets \cite{borisov2022deep}. In this work, we study three open and publicly packages that use different algorithms for finding CE. Since its release, the DiCE algorithm \cite{Mothilal:2019aa} has become one of the standard methods for CE. Now part of the Azure Responsible AI framework \cite{AzureCF} it is heavily used by data science practitioners. Optimization in general and DiCE in particular can be slow, and therefore workers in the field sought to take advantage of neighboring datapoints information. One such algorithm is NICE \cite{Brughmans:2021aa} which also allows for extensions that give more weight to plausibility if required. Lastly, in this work we also used the method proposed by Samoilescu et al. (2021) \cite{samoilescu2021model} that leverages deep reinforcement learning to learn the optimization procedure, thus avoiding the need to run the time-consuming optimization part during inference. We also expand on the foundational work of Russell (2019) \cite{Russell:2019aa}, by using Bayesian ML and investigate the use of the full posterior distribution in the optimization. 

Proximity-based CE in their current ML incarnation are usually an optimization process, which simply requires finding the minimum distance a data point has to cover in order to cross the decision boundary. Immediately however, a tension arises due to the ubiquitous categorical data that dominate most tabular datasets. Data science practitioners are trained to deal with categorical data by encoding them into binary columns which results to preprocessed datasets that are partly sparse and partly dense. Categorical data also cause complications in the optimization process for due to their discreetness defining a distance measure is challenging (Research Challenge 16 in Verma et al. (2020) \cite{Verma:2020aa}). 
The root of the complication is that categorical features are constrained in the form of data polytopes: for instance, a vehicle can belong to only one fuel type category—such as "electric"—which inherently excludes all others like "diesel," "petrol," or "hybrid." These mutually exclusive constraints define discrete regions in the feature space that valid data points must reside within. Furthermore, one naturally would expect some correlations between the features in the data: If a person doesn't own a car then they won't have car insurance. The hope is that explanation algorithms will learn this cross-column correlations, however it is important to note that not all datasets are perfect and noise can have a significant effect on the accuracy and matter-of-factness of those polytopes. Russell (2019) \cite{Russell:2019aa} used linear models and mixed integer linear programming (MILP) to achieve including categorical variables in the distance minimization. Therefore, this plausibility requirement, if not integral, it is a useful part of a CE.

Plausibility, in a more general sense however, can be deceptive as it can be understood at different levels \cite{keane2021if}. A standard example in the literature is that lowering one's age is not a realistic explanation for obtaining approval for a loan. At face value, this may not be immediately useful to the loan applicant, but for a stakeholder who wants to understand bias and fairness of the ML model it can be informative. Thus it would be advisable to data scientists to not make plausibility decisions unless the interests of all stakeholders are considered\footnote{Indeed, even informing the applicant that lowering their age would result in loan approval can be useful to them. Perhaps it will motivate them to ask deeper questions about the fairness of the system, or seek guidance on how to improve their chances through realistic and equitable means.}. 

Besides proximity and plausibility other desired properties of CE in the literature are sparsity and diversity. On the surface, sparsity is meant to ease the creation of actionable CE and can be achieved via including $\ell_0$ in the minimization process. However on the one hand this makes the process NP hard and on the other it is not always clear that hard sparsity should be encouraged. This can be problem-dependent. A resolution can be to minimize $\ell_1$, a soft sparsity constraint, which can be more desirable in general. On the other hand diversity of explanations can by accident lead to the Rashomon effect \cite{Anderson2016,Mehedi_Hasan_2022, Artelt:2022aa}, or by design to abuses of CE \cite{Krishna:2022aa,Brughmans2024}. It is thus important to have a clear understanding of how reliable CE are; and in order to make tangible progress in our analysis here, we focus on proximity only, constrained such that all variables are equally important.

In supervised classification, predictive performance is commonly described in terms of true positives, true negatives, false positives, and false negatives \cite{bishop2023deep}. For approximately balanced datasets, overall accuracy—defined as the proportion of correct predictions—serves as a widely used proxy for model performance. Importantly, accuracy is also affected by underlying sources of model uncertainty. In ML, uncertainty is often categorized as aleatoric (data-related noise and variability) and epistemic (stemming from limited data or model capacity) \cite{H_llermeier_2021, Gruber:2023aa}.

The predictive accuracy of the underlying ML model is, of course, a prerequisite for producing meaningful CE, since explanations of poor classifiers may themselves be unreliable. While predictive accuracy remains the dominant criterion for model selection, in explanation-driven applications it is equally important to assess the robustness of CE under uncertainty \cite{mishra2021survey,dominguez2022adversarial,black2021consistent,dutta2022robust,pawelczyk2020counterfactual,upadhyay2021towards,slack2021counterfactual,jiang2024robust,kavouras2024glance}. Notably, two models with similar predictive accuracy may yield CE that differ substantially in their stability. This motivates our investigation into how aleatoric and epistemic uncertainties affect counterfactual robustness, and why model selection based solely on predictive accuracy may be insufficient when explanations are part of the intended use case.

A very practical and timely use case of CE is in tabular datasets \cite{shwartz2022tabular, borisov2022deep} which frequently appear in social science and financial datasets \cite{grath2018interpretable, ustun2019actionable,karimi2021algorithmic,pawelczyk2021carla}. These datasets tend to have somewhat lower accuracies compared to other modalities such as text and image. They are subject to epistemic uncertainty e.g., due to incomplete knowledge of all input variables that affect the outcome. They are also subject to aleatoric uncertainty that can increase in time due to labeling errors, behavioural shifts, data collection drifts, environmental changes and sociocultural shifts in general\footnote{A practical example of this is the increase of gig workers and freelance economy in general. For these professions features like monthly salary and employment may not be as robust as they were in the past.}.  

Concrete examples of these uncertainties are abundant in tabular domains. For instance, epistemic uncertainty arises in credit scoring when relevant but unobserved variables—such as informal income streams or sudden changes in employment—are missing from the dataset \cite{louizos2017causaleffectinferencedeep}, leading the model to learn an incomplete decision boundary. Aleatoric uncertainty, by contrast, may result from mislabeled loan repayment records, evolving consumer behavior, or regulatory changes that shift the underlying data distribution \cite{hand1997statistical, song2022learningnoisylabelsdeep}. For machine learning models, such uncertainties reduce predictive reliability; for explainability methods, their consequences can be even more pronounced. CE depend directly on the learned decision boundary, and if this boundary is unstable or poorly supported by data, the resulting CE can shift drastically under small perturbations. This undermines their usefulness for end-users, who may receive recommendations that are not only difficult to act upon but also inconsistent over time. In high-stakes applications such as credit, employment, or healthcare, this lack of robustness threatens both user trust and the perceived fairness of the system \cite{slack2021counterfactual,Verma:2020aa,papanikou2025explanationsbiasdetectorscritical}.

Aside from the scientific interest for making use of CE in XAI, there is also the practical point of view of the data scientist practitioners and the considerations they have when choosing which XAI method to use. The time-constraints of the private and public organizations who are the end-users of these CE methods, coupled with the time this type of exploratory work takes, warrants a study comparing the performance of various CE methods in the presence of different sources of ML uncertainty. We hope this work will inform XAI research and serve as a guide to practitioners in equal measures.

We conclude this section by summarizing our main contributions herein:
\begin{itemize}

\item{We systematically quantify how different sources of ML uncertainty affect the robustness of proximity-based CE, using widely available models and CE libraries in a practitioner-friendly Python environment. Our evaluation pipeline mirrors real-world workflows, encompassing preprocessing, model training, deployment, and explanation generation. 
}

\item{We demonstrate that higher predictive accuracy of the ML classifier does not necessarily imply robustness in CE. This finding challenges the common practice of selecting models based solely on the classifier's predictive accuracy, when explanations are required, highlighting that accuracy does not automatically translate into more robust or stable CE. It also underscores the absence of a universally reliable combination of ML model and CE method for practitioners aiming for trustworthy explainability.}

\item{We find that minor accuracy drops—stemming from aleatoric uncertainty—can cause disproportionate shifts in individual and average CE, raising concerns about user trust and interpretability under uncertainty.}

\end{itemize}

In this study, we generated approximately 100,000 CE to analyze the behavior of model outputs across a diverse range of datasets, ML algorithms, and CE methods. All code and output data tables associated with this work are available on GitHub\footnote{\scriptsize{\url{https://github.com/leonidaschristodoulou/xai-cfs}}}.

This paper has the following structure: In Section~\ref{sec:related} we study previous works in the robustness of CE for tabular data, in Section~\ref{sec:method} we present our experimental and statistical methods and in Section~\ref{sec:results} our results. Section~\ref{sec:conclusions} discusses the limitations of our work and points to future research directions and lastly, Section~\ref{sec:conclusions} summarizes our findings.

\section{Related works and applications}
\label{sec:related}


CE robustness \cite{mishra2021survey,artelt2021evaluating,slack2021counterfactual,jiang2024robust,kavouras2024glance,dutta2022robust} has been identified as one of the central research challenges in explainability \cite{Verma:2020aa}. Different approaches have been proposed to probe this robustness. Slack et al. (2021) \cite{slack2021counterfactual}, for instance, perturbed the input of the CE algorithm to demonstrate instability, while others studied the impact of data distribution shifts and model drift \cite{rawal2020algorithmic,black2021consistent,meyer2024verified}, showing that drift can substantially degrade counterfactual reliability. In contrast, our analysis primarily uses controlled noise injection rather than distribution shifts. This choice enables us to directly modulate model uncertainty and accuracy in a systematic way, while avoiding the confounding factors introduced by real-world drift. By training multiple classifiers at varying noise levels, we can disentangle the specific role of aleatoric and epistemic uncertainty on counterfactual robustness. To our knowledge, this aspect has not been considered in prior work.

To our knowledge, the first publication that explicitly examined the impact of ML uncertainty on CE was Artelt et al. (2021) \cite{artelt2021evaluating}. Their study analyzed how CE are affected by noise, classifier choice, and CE algorithms, though it did not consider synthetic or real datasets with polytopes. They identified a “dimensionality curse,” where the $\ell$-norm difference between the original dataset and the dataset with added Gaussian noise grows with dimensionality. Artelt et al. also showed that plausible CE are more robust than those optimized solely on proximity, and argued that non-robustness can raise fairness concerns in certain contexts. While our focus is not on fairness per se, we extend this line of work by showing that even minor accuracy drops—stemming from aleatoric uncertainty—can cause disproportionate shifts in both individual and average CE, raising concerns for user trust and interpretability under uncertainty. 

In a recent study Brughmans et al. (2024) \cite{Brughmans2024} investigated the disagreement of CE, specifically how simply by using different CE algorithms one can get an array of different explanations. Clearly this is an undesirable property of CE since under the guise of XAI one can choose an explanation that suits them best, opening the gates for misuse \cite{goethals2023manipulation}. It is conceivable that the disagreement problem will escalate with the addition of different sources of ML uncertainty which is the main examination of this paper. We note that \cite{artelt2024effect} also study how CE can be exploited by malicious agents this time with data poisoning. 

To address the inevitable misclassifications made by ML classifiers, Asrzad et al. (2024) \cite{Asrzad24} introduced a distinction between CE for true and false predictions. Their approach identifies false classifications using trust scores \cite{Jiang:2018aa}, although this method has known limitations. Nonetheless, we agree with Asrzad et al. (2024) \cite{Asrzad24} that a fundamental limitation persists for simple CE that merely cross the decision boundary. Both false negatives and false positives have, by definition, already crossed the decision boundary—albeit in the incorrect direction. Therefore, offering a naive CE recommendation that suggests further movement in the same direction risks compounding the classifier’s error. In this work, we underscore this concern by analyzing the $\ell_1$ proximity between true negatives and false negatives in the context of CE.

Recent studies have proposed methods to mitigate robustness issues in counterfactual explanations and to offer guarantees under model perturbations. For instance, Jiang et al. (2024) \cite{jiang2024robust} introduce an interval-constrained optimization framework that explicitly accounts for model uncertainty when generating explanations. Other proposals integrate uncertainty quantification into CE generation: Delaney et al. (2021) \cite{delaney2021uncertainty} incorporate Bayesian uncertainty estimates into the optimization process. \cite{2024arXiv240804842S} propose a post-hoc method that provide model-agnostic probabilistic guarantees; Löfström et al. (2024) \cite{LOFSTROM2024123154} leverage predictive intervals to guide more reliable recourse recommendations; and Andringa et al. (2025) \cite{ANDRINGA2025102972} extend these ideas to longitudinal data settings. Together, these approaches highlight an emerging trend towards robustness-aware and uncertainty-aware CE. In contrast to these methods, our contribution is a comparative study across multiple datasets, ML algorithms, and CE methods, designed to help applied practitioners understand how different sources of uncertainty in ML models affect the robustness of counterfactual explanations.

\section{Methodology and Experimental Setup}
\label{sec:method}
\subsection{Experimental goals}
A central question in evaluating CE is their robustness: do the suggested CE remain stable when the underlying data or models are perturbed? In practice, CE are rarely computed on perfectly clean or noise-free data. Small fluctuations in the input distribution, or small changes in the trained model, may yield noticeably different counterfactuals. If such changes are large, the usefulness of the CE is undermined — because an end-user cannot trust that the explanation reflects a stable underlying mechanism.

To address this, we study how counterfactuals computed under increased noise deviate from their low-noise baseline counterparts. We quantify the difference using both absolute $\ell_1$	
distances and relative distances (i.e. distances scaled by the baseline CE norm), which provide interpretable, scale-free measures of robustness. The goal of this analysis is not to test whether differences exist (they inevitably will), but to characterize how much CE change as noise increases, and whether those changes remain within acceptable bounds.

\subsection{Data and Methods}

We conduct our numerical experiments with synthetic data using the \texttt{make\_classification} function from \texttt{scikit{-}learn} \cite{scikit-learn}, a data generation tool commonly used for benchmarking and evaluating classification algorithms. \texttt{make\_classification} generates synthetic classification datasets by sampling points from Gaussian-distributed clusters centered around class-specific means. It allows for fine-grained control over feature redundancy, noise and class imbalances, making it a flexible tool for testing ML models. In this paper we use it for testing the relation of CE and the accuracy of ML models.
 All the synthetic datasets that we used are shown in Table \ref{tab:mock_characteristics}. We created polytopes by discretizing continuous variables. 

We also analyze three real-world datasets - German Credit, Adult Income and Give Me Some Credit
- with characteristics that are particularly relevant to XAI. Loan applications and salary estimation are among the most common use cases for counterfactual explanations. These datasets are all used for binary classification and are complementary, as they vary in size and feature composition, covering a broad range of scenarios, making them representative for this study.

 \begin{table}[t]
    \centering
    \small
    \resizebox{\columnwidth}{!}{%
    \begin{tabular}{crrrrrrr}
        \toprule
        Mock & N samples & N features & N informative & N categorical & N Polytopes & Non-iid & Missing Variables \\
        \midrule
        \hline
        1 & 3000 & 2 & 2 & 0 & 0 & F & F \\
        2 & 3000 & 12 & 10 & 0 & 0 & F & F \\
        3 & 3000 & 10 & 10 & 5 & 2 & F & F \\
        4 & 5000 & 40 & 40 & 20 & 10 & F & F \\
        5 & 3000 & 10 & 10 & 5 & 2 & T & F \\
        6 & 3000 & 7 & 7 & 4 & 1 & F & T \\
        \bottomrule
    \end{tabular}%
    }
    \caption{The characteristics of the mocks datasets we constructed for investigating the impact of aleatoric uncertainty to counterfactual explanations.}
    \label{tab:mock_characteristics}
\end{table}

\begin{figure*}[t]
    \centering
    \subfigure[]{
        \includegraphics[width=0.28\textwidth]{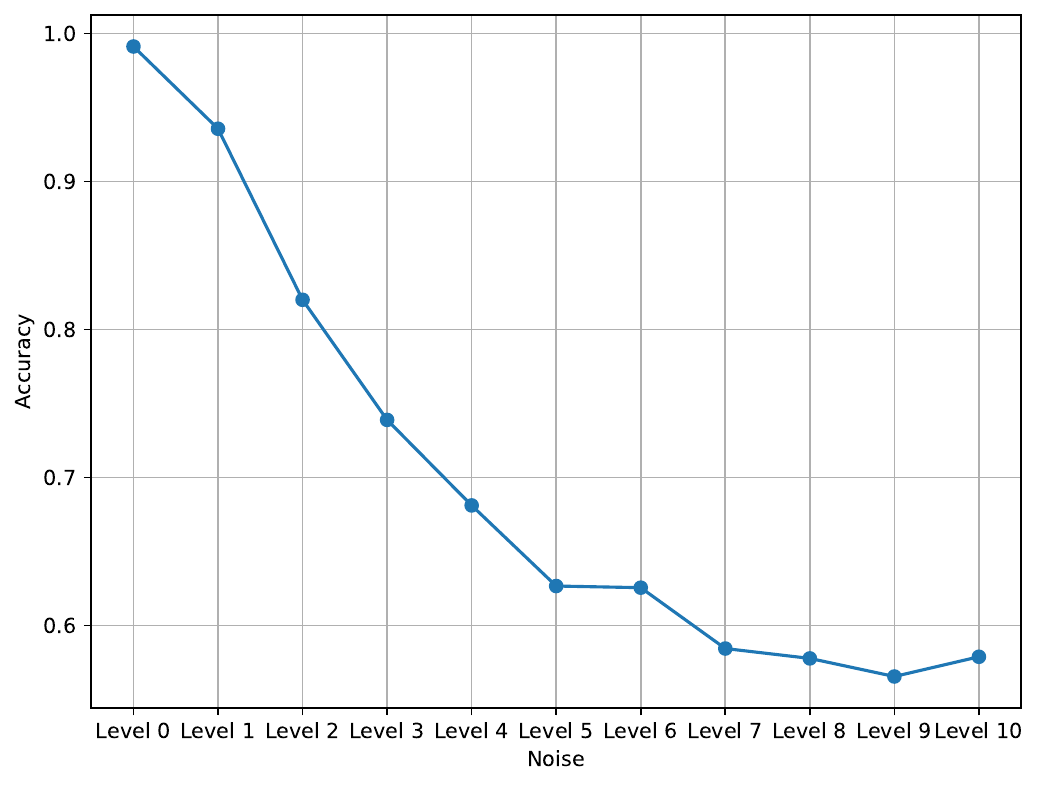}
        \label{fig:mock_accuracy_example1}
    }
    \subfigure[]{
        \includegraphics[width=0.66\textwidth]{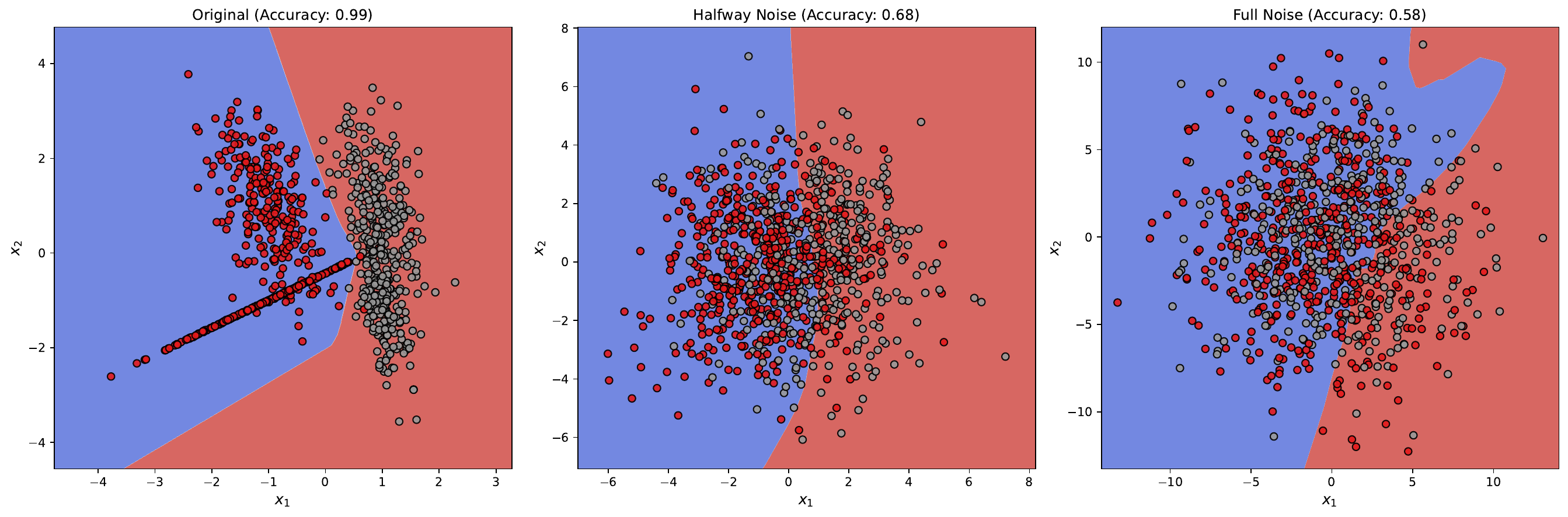}
        \label{fig:decision_boundary}
    }
    \caption{(a): The impact of noise level in model accuracy for mock dataset 1. The uncertainty added in this example is purely aleatoric (b): The impact of noise on the model's decision boundary for dataset 1. As the noise increases the decision boundary becomes more complicated and consequently it increases the error on the CE.}
    \label{fig:mock_accuracy_combined}
\end{figure*}
\subsection{ML Classifiers}
We chose to work with binary classification, a very standard applied data science problem. The mock datasets are relatively balanced ($\approx60\%$). German Credit ($30\%$) and Adult Income datasets ($24\%$) are moderately imbalanced, whereas The Give Me Some Credit dataset is strongly imbalanced ($93\%$ negative, $7\%$ positive). We emphasize that our study does not aim to benchmark predictive performance; instead, accuracy is used as a coarse control for the level of noise-induced degradation in the models. However, since target imbalance is important for stakeholders, we also present the CE robustness analysis split by TN and FN.  

We use 4 different classification algorithms: regularized logistic regression (LR), bayesian logistic regression (BLR), random forests (RF) and neural networks (NN). 

\subsection{Counterfactual Generation Methods}

\begin{table}[t]
    \centering
    \resizebox{\columnwidth}{!}{%
    \begin{tabular}{l|l|l|l|l}
    \hline
    Classification Model & Library    & ML Reference & CE Method        & CE Reference \\ \hline
    \hline
    Regularised Logistic Regression    & \texttt{scikit-learn}   & \cite{scikit-learn}       & \texttt{MILP}               & \cite{Russell:2019aa}       \\ 
          &            &           & DiCE               & \cite{Mothilal:2019aa}       \\ 
          &            &           & NICE               & \cite{Brughmans:2021aa}       \\ 
          &            &           & RL                 & \cite{samoilescu2021model}       \\ \hline
    Bayesian Logistic Regression    & \texttt{PYMC}       & \cite{pymc2023}        & \texttt{MILP (mean)}        & \cite{Russell:2019aa}       \\ 
          &            &           & \texttt{MILP (marg)}        & \cite{Russell:2019aa}      \\ \hline
    Random Forests    & \texttt{scikit-learn}   & \cite{scikit-learn}       & DiCE               & \cite{Mothilal:2019aa}       \\ 
          &            &           & NICE             & \cite{Brughmans:2021aa}       \\ 
          &            &           & RL                 & \cite{samoilescu2021model}     \\ \hline
    Neural Networks    & \texttt{TensorFlow}         & \cite{tensorflow2015-whitepaper}       & DiCE               & \cite{Mothilal:2019aa}       \\ 
          &            &           & NICE               & \cite{Brughmans:2021aa}       \\ \hline
    \end{tabular}
    }
    \caption{The classification algorithms and CE methods we used throughout this paper.}
    \label{tab:algorithm_libraries}
\end{table}

We obtain CE using a modified version of Russell (2019) MILP method \cite{Russell:2019aa}, DiCE, NICE and \cite{samoilescu2021model} RL method. We note that the MILP method is a generalization of the Wachter et al. CE method with more stable solutions to the optimization problem and the ability to work with integer constraints. Not all CE methods can easily be applied to all classifiers
\footnote{Throughout this paper we have used a consistent Python environment (available on \texttt{GitHub}), to ensure reproducibility and a fair comparison across methods. This setup enables unified data preprocessing and model handling, minimizing potential inconsistencies. However, it also introduced version constraints: for example, the RL-based counterfactual method requires \texttt{TensorFlow 2} version $\leq 2.15$. Other CE methods, like Counterfactuals Guided by Prototypes \cite{van2021interpretable} and Contrastive Explanations \cite{dhurandhar2018explanations}, require disabling \texttt{TensorFlow 2} behaviour(\texttt{tf.compat.v1.disable\_v2\_behavior()}), as it relies on legacy \texttt{TensorFlow 1} constructs. Consequently, we could not include them in the analysis, since they are not compatible with current \texttt{TensorFlow 2} environments without major reimplementation. Finally, we note that the gradient descent option of DiCE was not working in our python environment, so we reverted to random search. These issues do not affect our conclusions but they do highlight limitations that practitioners face in the field.}; 
our final combinations are also shown in Table \ref{tab:algorithm_libraries}. 

Polytopes are probably the most important feature that distinguishes categorical datasets from other data modalities. The CE methods we employ handle polytopes in different ways. The MILP
\footnote{We use the Gurobi library in our MILP implementation \cite{gurobi}} 
method searches for solutions that satisfy the constraints imposed by each polytope. In our implementation, we also allow for dimensionality reduction when using one-hot-encoding, as one column can be dropped due to redundancy. This is a common course of action in applied data science. Consequently, in our LR + MILP approach, each polytope can sum to either 0 or 1.

Maintaining consistency whilst testing different CE methods is challenging in practice, due to each method accepting inputs at a different preprocessing stage. LR + MILP require the completely processed dataset whereas DiCE, NICE (when using \texttt{scikit-learn}) and the RL method required the preprocessed dataset with the preprocessing functions. On the other hand, when using \texttt{TensorFlow}, both DiCE and NICE require the processed dataset with the one-hot-encoding. Note however, that all methods require manual entering of the categorical feature names in some fashion. 

\subsection{Robustness Metrics}
We want to investigate how much a CE changes as the underlying ML model loses accuracy. To this end we need to define the $\ell_1$ difference between two vectors $\mathbf{a}$ and $\mathbf{b}$ in mixed space with continuous and categorical dimensions \cite{Russell:2019aa}. Let $\mathbf{a}, \mathbf{b} \in \mathbb{R}^d$ be two CE data points with:
\begin{itemize}
    \item $\mathcal{C}_{\text{cat}} \subset \{1, \dots, d\}$: indices of categorical features
    \item $\mathcal{C}_{\text{cont}} \subset \{1, \dots, d\}$: indices of continuous features
\end{itemize}
Let be the weight vector $\mathbf{w} = (w_1,...,w_j)$ defined as:
\[
w_j = 
\begin{cases}
\displaystyle\frac{1}{\Phi^{-1}(0.75)\tilde{\sigma}_j}, & \text{if } j \in \mathcal{C}_{\text{cat}} \\[10pt]
\displaystyle\frac{1}{\tilde{\text{MAD}}_j}, & \text{if } j \in \mathcal{C}_{\text{cont}}
\end{cases}
\]
where $\Phi$ is the cumulative distribution function for the normal distribution. For continuous variables MAD is defined as
\begin{align*}
\text{MAD}_j &= \operatorname{median}_{i=1}^n \left( \left| x_{ij} - \operatorname{median}_{i=1}^n(x_{ij}) \right| \right), \quad j \in \mathcal{C}_{\text{cont}}
\end{align*}
and for categorical variables $\sigma_j$ is defined as
\begin{align*}
\sigma_j &= \sqrt{ \frac{1}{n} \sum_{i=1}^n \left( x_{ij} - \bar{x}_j \right)^2 }, \quad j \in \mathcal{C}_{\text{cat}}
\end{align*}
The overall weighted $\ell_1$ distance between $\mathbf{a}$ and $\mathbf{b}$ is defined as:
\begin{equation}
D(\mathbf{a}, \mathbf{b}) = \sum_{j=1}^d w_j \left| a_j - b_j \right|
\label{eq:distance}
\end{equation}
Eq. \ref{eq:distance} can easily be extended for other $\ell-norms$. For the remainder of the paper we use the the normalised distance
\begin{equation}
D_{\text{rel}}(\mathbf{a}, \mathbf{b}) = \frac{\sum_{j=1}^d w_j \left| a_j - b_j \right|}{{\sum_{j=1}^d w_j \left|b_j \right|}}
\label{eq:distance_norm}
\end{equation}
which allows us to compare directly datasets with different dimensionalities. 

We evaluate the robustness of counterfactual explanations (CE) by measuring how CE generated under increased uncertainty deviate from their baseline counterparts. From $D_{\text{rel}}(\mathbf{a}, \mathbf{b})$, we derive three complementary robustness metrics:
\begin{itemize}
\item{Central robustness. To capture the typical magnitude of change, we report the median relative distance along with bootstrap confidence intervals. This provides a scale-free summary of how far CE drift under uncertainty.}
\item{Distributional robustness. To assess variability across instances, we report the median, the 10th and 90th percentiles, and the interquartile range (IQR) of the robustness distribution. These statistics capture whether robustness holds consistently across the population, rather than only on average.}
\item{Comparative robustness. To compare methods or models, we evaluate differences in medians, compute standardized effect sizes, and perform a paired non-parametric test (Wilcoxon signed-rank test). This analysis identifies whether certain approaches yield systematically more stable CE under uncertainty.}
\end{itemize}

Together, these metrics provide complementary views of robustness: central tendency (median drift), distributional spread (tail risk and tolerance rates), and comparative sensitivity across models.

\section{Results}
\label{sec:results}
\subsection{Synthetic data results}


\begin{figure}[t]
  \centering
  \includegraphics[width=\columnwidth]{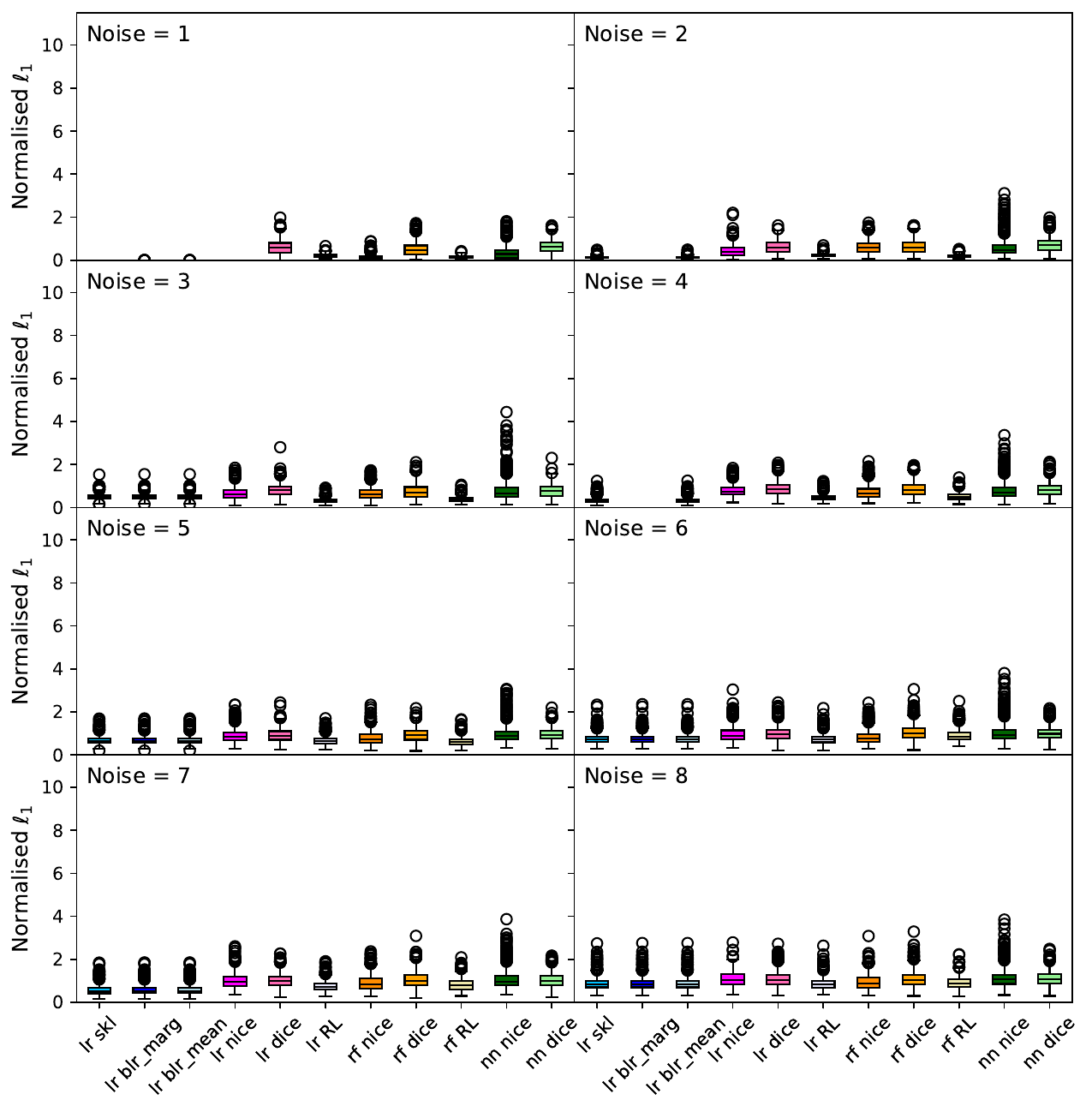}
  \caption{%
    The distribution of the normalised $\ell_1$–difference of the CE from the ground noise level,
    for mock dataset 2 (10 continuous predictive features). Each box spans Q1–Q3, with a
    line at the median; whiskers extend to 1.5 times IQR, and outliers beyond are shown as circles.%
  }
  \label{fig:l1_2_page}
\end{figure}

\subsubsection{Aleatoric Uncertainty}
\label{sec:aleatoric}

Aleatoric uncertainty arises from inherent noise in the data—randomness that cannot be eliminated, even with more information. In our synthetic data experiments, we introduce such noise in two controlled ways: (1) by adding Gaussian noise to the input features, and (2) by randomly flipping a fraction of the target labels. These modifications allow us to simulate different noise levels, designed to produce an approximately linear degradation in model accuracy. Figure~\ref{fig:mock_accuracy_example1} illustrates how we regulate accuracy by varying the noise level in one representative case (mock dataset 4). Figure~\ref{fig:decision_boundary} further demonstrates how increasing noise distorts the learned decision boundaries (for mock dataset 1), leading to a progressive decline in classification performance.

We replicate the datasets in Table \ref{tab:mock_characteristics} and for each replication we add a different noise level. We split each dataset to a training and test set with 30\% ratio. Rather than using cross-validation, we probe robustness by systematically injecting uncertainty in the data. We then run the classification and CE algorithms according to Table \ref{tab:algorithm_libraries}. We treat the lowest noise level as our ground truth and then measure the normalized $\ell_1$ distance of each higher noise level CE with respect to that. The closest to the noise level 0 the CE is the better.

\begin{figure}[t]
  \centering
  \includegraphics[width=\columnwidth]{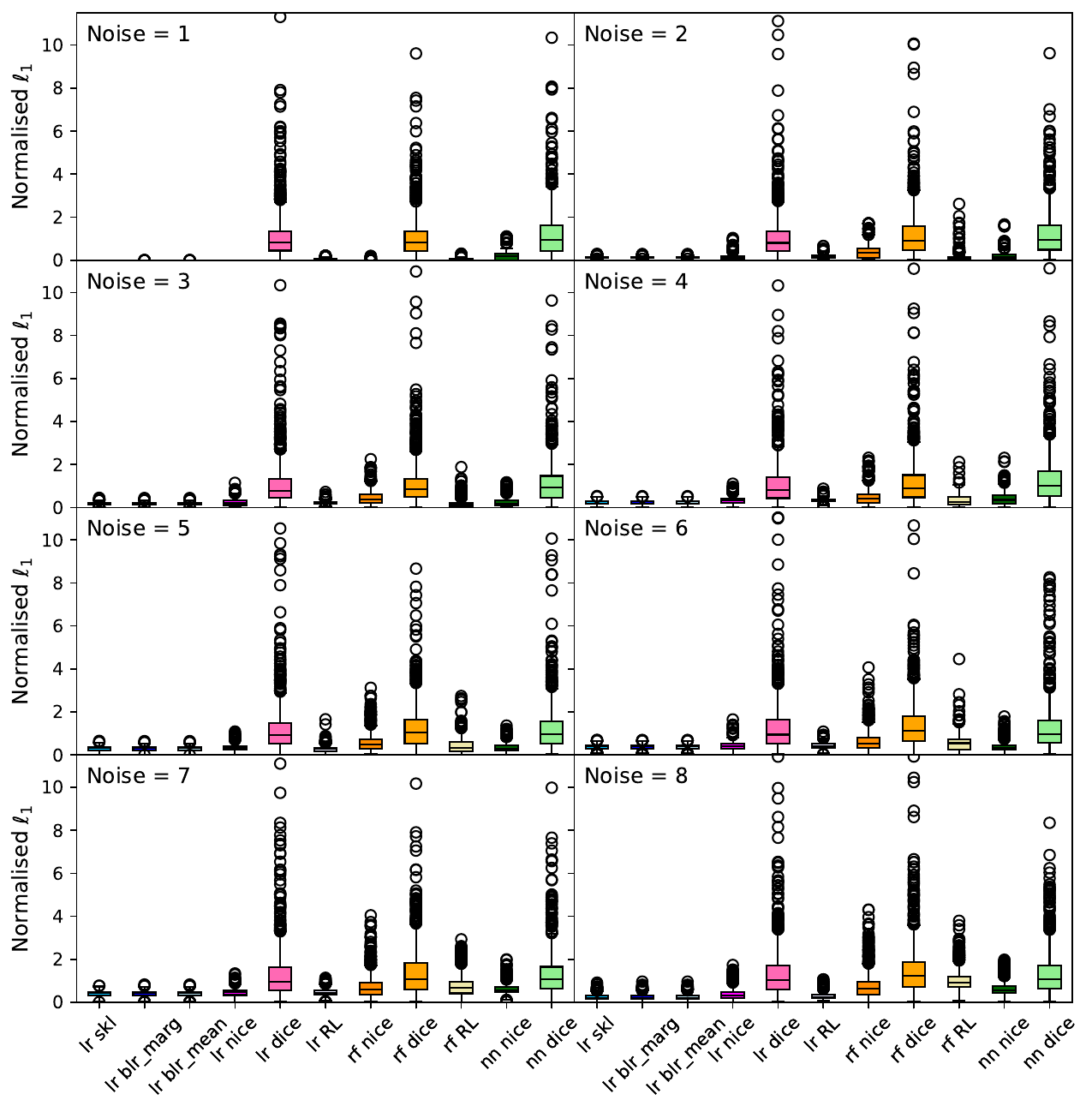}
  \caption{%
    The distribution of the normalised $\ell_1$–difference of the CE from the ground noise level,
    for mock dataset 1 (2 continuous predictive features). Each box spans Q1–Q3, with a
    line at the median; whiskers extend to 1.5 times IQR, and outliers beyond are shown as circles.%
    }
  \label{fig:l1_1_page}
\end{figure}

\begin{figure}[H]
    \centering
    \subfigure[]{
        \includegraphics[width=0.45\columnwidth]{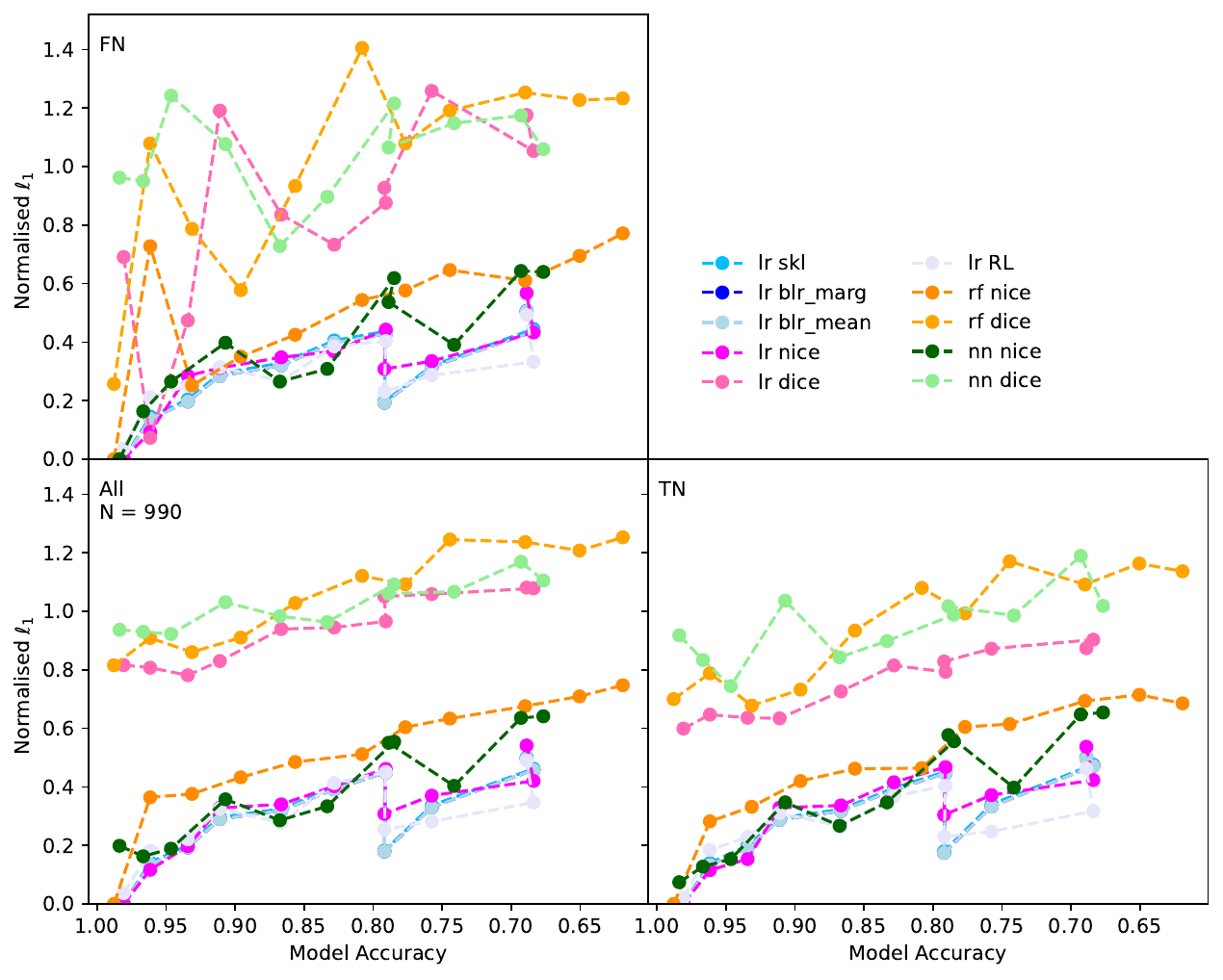}
        \label{fig:accuracy_1}
    }
    \subfigure[]{
        \includegraphics[width=0.45\columnwidth]{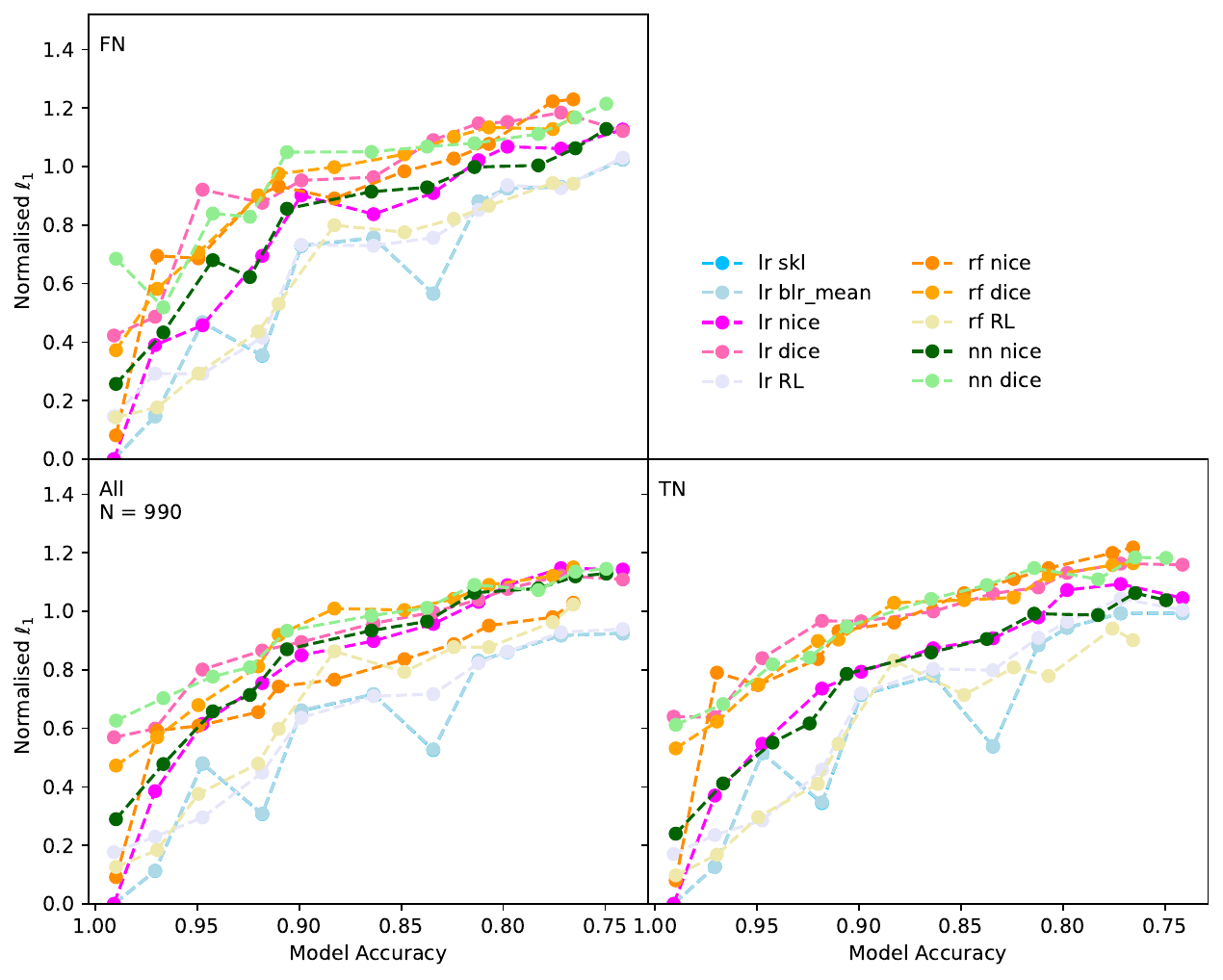}
        \label{fig:accuracy_2}
    }
    \subfigure[]{
        \includegraphics[width=0.45\columnwidth]{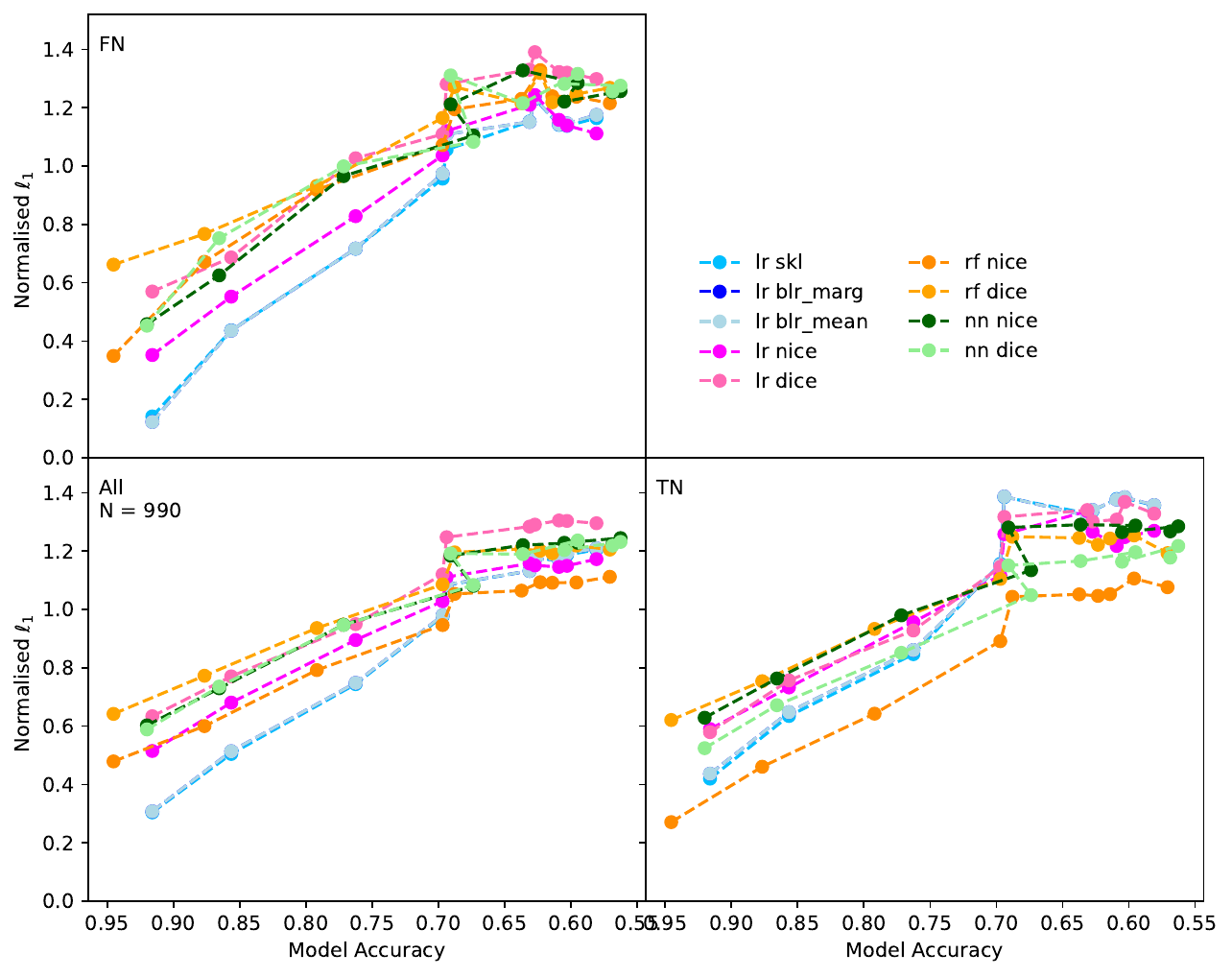}
        \label{fig:accuracy_3}
    }
    \subfigure[]{
        \includegraphics[width=0.45\columnwidth]{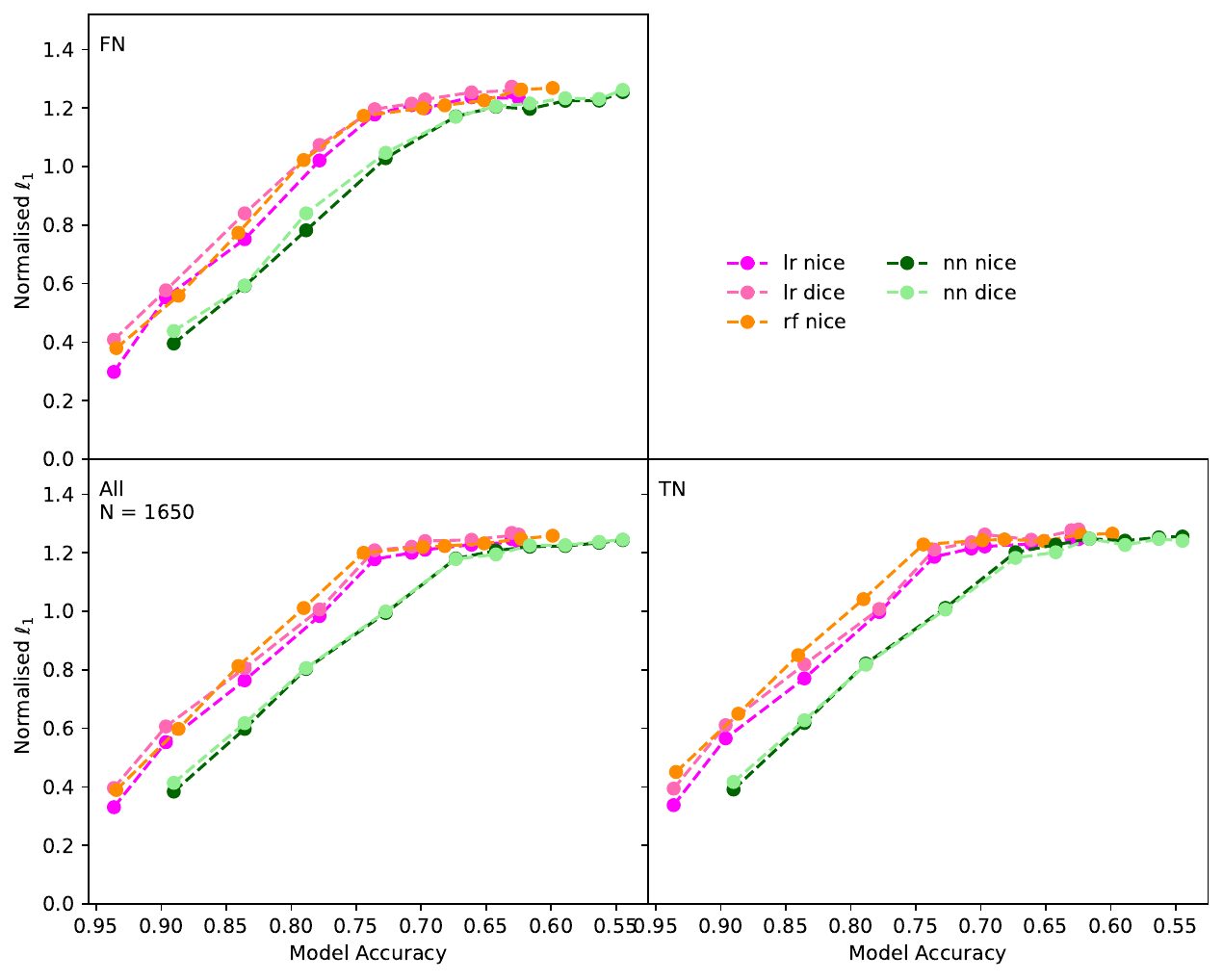}
        \label{fig:accuracy_4}
    }
    \caption{$\ell_1$ distance difference from the ground state with minimal noise for various combinations of classification algorithms and CE methods again model accuracy. Only combinations with CE completeness $>10\%$ are included. Each panel shows the CE of test set of the classification exercise split into FN, all and TN.}
    \label{fig:accuracy_combined}
\end{figure}

Fig. \ref{fig:l1_1_page} and to some extend Fig. \ref{fig:l1_2_page} show the desired scenario from most stakeholders where the distribution of CE remains relatively stable as the noise increases. Despite the ever increasing noise these are ideal datasets, albeit without categorical variables, that help to set the stage for the behaviour of CE methods under uncertainty. 

In Fig. \ref{fig:accuracy_combined} we move on from noise levels and we show the relationship of the median normalised $\ell_1$ distance against model accuracy, which is the usual observable in applied data science. We also present results for False Negatives (FN) and True Negatives (TN) as another proxy of the CE robustness. This is distinction highlights tensions between different stakeholders. For high-stakes decision-making—such as job applicant screening—the distinction between FN and TN plays a critical role in the interpretation and utility of CE. FN represent missed opportunities: qualified individuals incorrectly rejected by the model. For these individuals, CE serve as a crucial tool for transparency and empowerment, offering concrete and actionable feedback on what minimal changes could have altered the model's decision. 
On the other hand, TN, while seemingly unproblematic from a performance metrics perspective, also benefit from counterfactual insights. For stakeholders such as HR professionals, compliance officers, and auditors, CE for TN cases help justify decisions, ensure procedural fairness, and reveal potential structural biases. If a subgroup consistently receives CE that are unrealistic or unattainable, this can signal deeper issues with model bias or representational gaps in training data. Therefore, analyzing both FN and TN CE is essential—not only to improve individual outcomes but also to promote system-level transparency, trust, and equity.

We reverse the x-axis of Fig. \ref{fig:accuracy_combined} in order to keep the best possible outcome close the axes origin. If there is large number ($>10\%$) of CE that are wrong, either because an optimizer didn't find a solution or because the CE is wrong, then the data for the ML-CE combination is left out. Panels \ref{fig:accuracy_1}, \ref{fig:accuracy_2}, \ref{fig:accuracy_3} and \ref{fig:accuracy_4} show the $\ell_1$ distance vs model accuracy for mock datasets 1, 2, 3 and 4 respectively. We observe that the $\ell_1$ increases as the complexity of the dataset increases but also that there exist many differences between the various ML and CE algorithm combinations. 

First we remark that for the frequentist and bayesian LR the MILP CE results are very similar and that we don't find significant gain by using the full posterior for the optimization. On the other hand we find that CE from DiCE and NICE have accuracy that scales somewhat worse with noise for linear ML models. TN and FN CE are very similar for LR classifiers. 

The increase in the $\ell_1$ percentage difference in Fig. \ref{fig:accuracy_combined} is disproportionate to the percentage decrease in model accuracy across all models. Since both quantities are dimensionless, this indicates that the rate of change for counterfactual error differs from that of model accuracy. This effect is particularly pronounced for datasets 3 and 4, which contain a larger number of features (see Table \ref{tab:mock_characteristics}). Interestingly, as the accuracy drops to around 0.7, the $\ell_1$ values begin to plateau, suggesting that the CE stabilize. However, before this point, a 25\% decrease in accuracy results in a doubling or even tripling of the $\ell_1$ values for mock datasets 3 and 4, respectively. This behaviour is also evident in real world datasets as will see in Section \ref{sec:real}.

RF are among the most popular classifiers for tabular data due to their inherent interpretability, accuracy and robustness. We create CE from RF using DiCE, NICE and RL, and for our mock datasets we find that they behave very differently with respect to $\ell_1$ even for the two-feature dataset, with NICE having the smaller $\ell_1$ in 3 out of 4 mock datasets, where RL performs best in the remaining one. We note that the CE from DiCE didn't pass our completeness threshold for mock datasets 1 and 2 and RL similarly for datasets 3 and 4. To sum up, for RF we find that three of the most widely used and publicly available CE methods output CE that are in discrepancy. The splitting by TN and FN however, does not reveal any significant differences, apart from dataset 3 where RF nice performs best only for the TN subset.  

Lastly, NN perform with average accuracy for datasets with small number of features, but rise to the top for the dataset with largest number of features and samples. There are not big variations between our two CE methods that compare (DiCE and NICE), although we note that for DiCE we used the random search optimizer. We note however that for datasets 1 and 3 NICE CE have different relative accuracy for FN and TN compared to other CE methods. DiCE CE are also unstable in this split, for dataset 1 they most robust, but they become last for dataset 3.  

To assess the robustness of CE methods across different model types, we evaluated per-instance changes in the generated explanations under increasing levels of input noise. We focused on three representative noise levels — Low, Medium, and High and for each method, we computed the per-instance distances between the noisy and baseline explanations, using a normalized $\ell_1$ metric that captures the overall magnitude of the perturbation in a scale-invariant way.

From these metric, we derived a set of descriptive statistics to characterize the robustness of each method. Specifically, we computed the median, the 10th and 90th percentiles, and the interquartile range (IQR) for each method and noise level. The median provides a measure of central robustness, indicating the typical deviation from the baseline, while the percentiles and IQR capture the distributional robustness, reflecting the variability across CE instances. These summary statistics allow us to compare methods in a way that is independent of the scale of the features or the underlying model accuracy. For clarity, hereafter we present one representative table in the main text, while the remaining tables for the other datasets are provided in~\ref{sec:appendix}.

Tables \ref{table:descr1},~\ref{table:descr2},~\ref{table:descr3} and~\ref{table:descr4} show the results of the descriptive statistics, highlighting the best performing ML-CF method for each statistic. Notably for all synthetic datasets there isn't a CE combination that dominates as the best performing one. Their performance varies and is dependent on the level of uncertainty and the statistic. 

\begin{table}[t]
\centering
\resizebox{\columnwidth}{!}{%
\begin{tabular}{lllllllllllll}
\toprule
 & \multicolumn{3}{c}{Median} & \multicolumn{3}{c}{P10} & \multicolumn{3}{c}{P90} & \multicolumn{3}{c}{IQR} \\
Uncertainty & High & Low & Medium & High & Low & Medium & High & Low & Medium & High & Low & Medium \\
ML-CF Method &  &  &  &  &  &  &  &  &  &  &  &  \\
\midrule
lr-RL & 0.25 & 0.04 & 0.33 & 0.16 & 0.01 & 0.21 & 0.49 & 0.11 & 0.45 & 0.18 & 0.06 & \textbf{0.11} \\
lr-blr (marg) & \textbf{0.18} & 0.00 & 0.29 & 0.09 & 0.00 & 0.10 & \textbf{0.43} & 0.00 & 0.35 & 0.16 & 0.00 & 0.12 \\
lr-blr (mean) & 0.18 & 0.00 & 0.29 & 0.09 & 0.00 & 0.10 & 0.43 & 0.00 & \textbf{0.35} & 0.16 & 0.00 & 0.12 \\
lr-dice & 1.05 & 0.82 & 0.83 & 0.31 & 0.20 & 0.23 & 2.70 & 2.16 & 2.23 & 1.11 & 0.90 & 0.97 \\
lr-nice & 0.31 & \textbf{0.00} & 0.33 & 0.12 & \textbf{0.00} & 0.13 & 0.66 & \textbf{0.00} & 0.49 & 0.28 & \textbf{0.00} & 0.19 \\
lr-skl & 0.18 & \textbf{0.00} & 0.30 & \textbf{0.09} & \textbf{0.00} & 0.11 & 0.43 & \textbf{0.00} & 0.36 & \textbf{0.16} & \textbf{0.00} & 0.12 \\
rf-RL & 0.92 & 0.05 & \textbf{0.24} & 0.50 & 0.02 & \textbf{0.07} & 1.66 & 0.13 & 0.67 & 0.48 & 0.04 & 0.37 \\
rf-dice & 1.25 & 0.82 & 0.91 & 0.41 & 0.18 & 0.27 & 2.68 & 2.05 & 2.31 & 1.16 & 0.92 & 1.04 \\
rf-nice & 0.63 & \textbf{0.00} & 0.43 & 0.21 & \textbf{0.00} & 0.13 & 1.47 & \textbf{0.00} & 0.91 & 0.57 & \textbf{0.00} & 0.41 \\
nn-dice & 1.06 & 0.94 & 1.03 & 0.34 & 0.18 & 0.30 & 2.51 & 2.31 & 2.44 & 1.09 & 1.20 & 1.14 \\
nn-nice & 0.55 & 0.20 & 0.36 & 0.30 & \textbf{0.00} & 0.10 & 1.14 & 0.36 & 0.74 & 0.31 & 0.31 & 0.37 \\
\bottomrule
\end{tabular}
}
\caption{CE descriptive statistics for mock dataset 1. The best performing ML-CF method of each statistic is highlighted.}
\label{table:descr1}
\end{table}

\begin{table}[ht]
\centering
\resizebox{\columnwidth}{!}{%
\begin{tabular}{lllllll}
\toprule
 & \multicolumn{2}{c}{Median Delta} & \multicolumn{2}{c}{Posterior P(best)} & \multicolumn{2}{c}{Significance} \\
Uncertainty & High & Medium & High & Medium & High & Medium \\
Method &  &  &  &  &  &  \\
\midrule
lr-RL & -0.07 & -0.05 & 1.0 & 1.0 & *** &  \\
lr-blr (marg) & NaN & 0.01 & NaN & 0.22 & NaN &  \\
lr-blr (mean) & -0.00 & 0.01 & 0.50 & 0.22 & *** &  \\
lr-dice & -0.87 & -0.57 & 1.0 & 1.0 & *** &  \\
lr-nice & -0.12 & -0.04 & 1.0 & 1.0 & *** &  \\
lr-skl & -0.00 & -0.00 & 0.74 & 0.50 & *** &  \\
rf-RL & -0.74 & NaN & 1.0 & NaN &  & NaN \\
rf-dice & -1.04 & -0.64 & 1.0 & 1.0 &  &  \\
rf-nice & -0.44 & -0.15 & 1.0 & 1.0 & *** &  \\
nn-dice & -0.90 & -0.76 & 1.0 & 1.0 & *** &  \\
nn-nice & -0.37 & -0.08 & 1.0 & 1.0 & *** &  \\
\bottomrule
\end{tabular}
}
\caption{Significance tests and posterior probabilities for method robustness under different levels of uncertainty for mock dataset 1. The left block reports mean robustness differences relative to the baseline, with asterisks marking significance levels ($*** p<0.001$, $** p<0.01$, $* p<0.05$). The right block reports the posterior probability that each method is the best-performing one among all candidates. NaN indicates cases where the test was not defined due to instability. “Posterior P(best)” denotes the probability, conditional on the observed data, that the given method is the top-performing one among the set of candidates.}
\label{table:p1}
\end{table}

To formally assess the statistical significance of the observed differences, we conducted two complementary tests. First, we used the Wilcoxon signed-rank test to determine whether the per-instance distances of a given ML-CE combination were systematically larger than those of the most robust method at each noise level. Second, we implemented a Bayesian comparison using a Student-T model to estimate the posterior distribution of the mean difference in per-instance distances between the best and each alternative ML-CE combination. This approach yields the posterior probability that the selected best method is truly the one with the smaller distance from the base level, providing a probabilistic measure that is both interpretable and robust to outliers (due to the use of the Student-T model for the likelihood).
Together, these analyses allow us to quantify and compare robustness across methods and noise levels, ensuring that observed differences are meaningful and not driven by a few extreme instances. 

Tables~\ref{table:p1},~\ref{table:p2},~\ref{table:p3} and~\ref{table:p4} report both the frequentist significance tests and the Bayesian posterior probabilities of being the best method across three uncertainty levels. Negative values in the significance columns indicate the average difference in robustness relative to the baseline, with asterisks marking statistical significance levels ($*** p<0.001$, $** p<0.01$, $* p<0.05$). The posterior probabilities complement these results by quantifying the likelihood that a method is the best-performing one among all candidates, given the observed data.

Across all four datasets, the results from both the frequentist significance tests and the Bayesian posterior probabilities are consistent. In nearly all cases, posterior probabilities are very close to 1, indicating strong support that the identified best-performing methods are indeed superior. The significance tests largely agree, with highly significant p-values, and only a few isolated cases producing undefined values due to instability. Taken together, these complementary analyses add weight to our hypothesis that the observed differences in CE robustness across methods are statistically significant and not artifacts of noise or sampling.

In conclusion, due to the presence of aleatoric uncertainty we find that no combination of classifier and CE method can guarantee best performance for CE based on proximity. There is no obvious recipe for data science practitioners that they can use and be confident that their CE are the best possible ones.

\begin{figure}[h]
\centering
\includegraphics[width=\columnwidth]{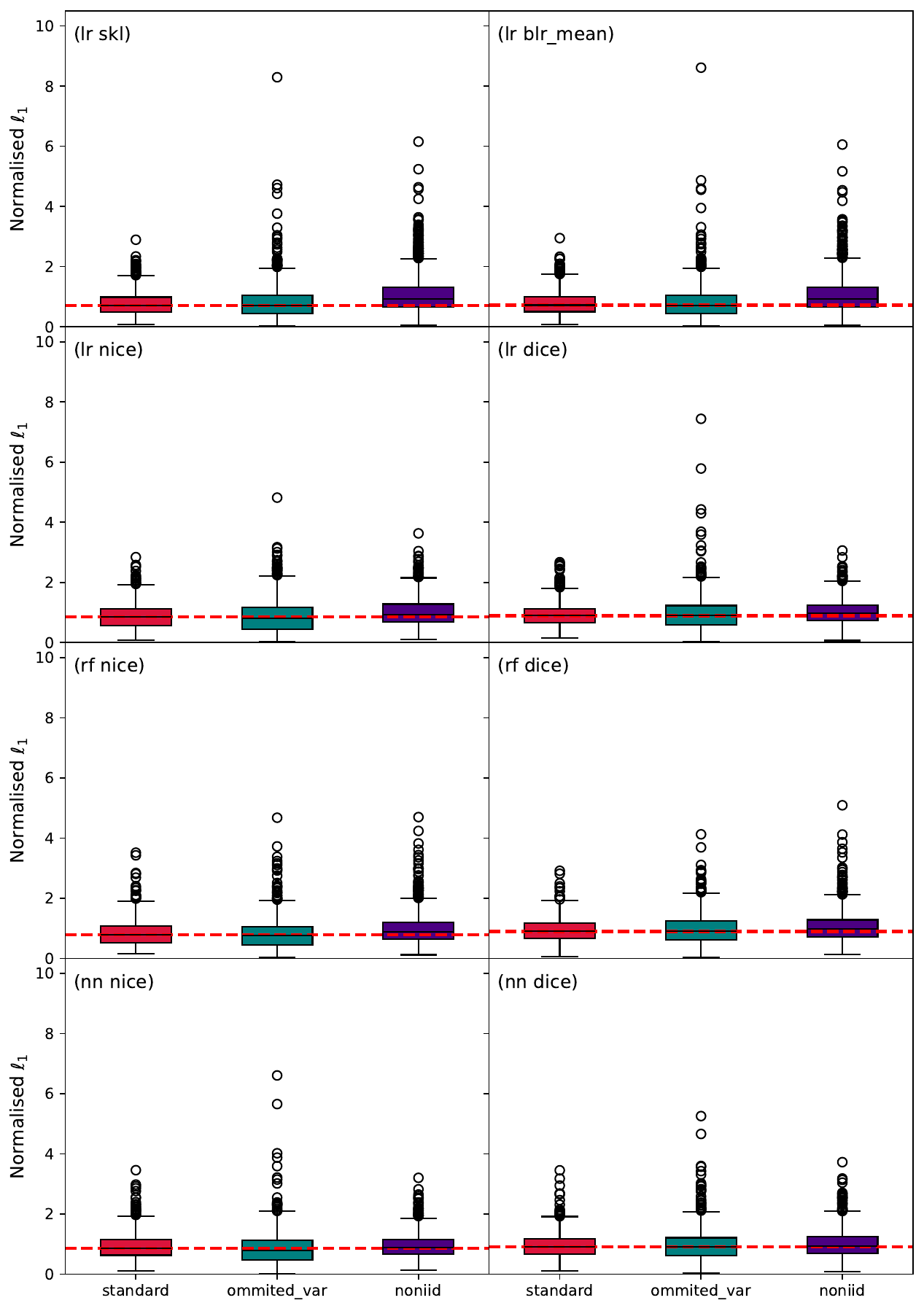}
\caption{The distribution of the normalized $\ell_1$ norm for different ML and CE methods in the presence of both aleatoric and epistemic uncertainty. The construction of the boxes, whiskers and outliers is the same as in Fig. \ref{fig:l1_1_page}. The dashed red line depicts the baseline median CE distance for the case with aleatoric uncertainty present only.}
\label{fig:epistemic_l1}
\end{figure}

\subsubsection{Epistemic Uncertainty}
\label{sec:epistemic}


In most applied data science instances our knowledge is incomplete far beyond the presence of random noise. Slightly simplifying, epistemic uncertainty encompasses the sources of ML uncertainty that results from unknown modelling errors, non-gaussian errors or a misspecified ML model. 

In this subsection we work solely with mock dataset 3 from the previous section. We introduce epistemic uncertainty in two ways: First by introducing non gaussian noise, indicating a possible misunderstanding in the data generation process and second by omitting some numerical variables demonstrating a lack of knowledge about the true input features. For every dataset we progressively add noise like in Section \ref{sec:aleatoric}. We then compare using Eq.\ref{eq:distance} how the distribution of $\ell_1$ changes. 

In Fig. \ref{fig:epistemic_l1} we compare the robustness of various combinations of ML and CE methods against joint aleatoric and epistemic uncertainty. The CE estimated are mostly robust, with only a small distribution shift observed for non-iid errors for linear models, and RF. Notably the CE for the NN classifier exhibit the smaller, close to zero,  distribution shift.

\subsection{Real-world Data Results}
\label{sec:real}

\subsubsection{German Credit}

\begin{figure}[h]
\centering
\includegraphics[width=\columnwidth]{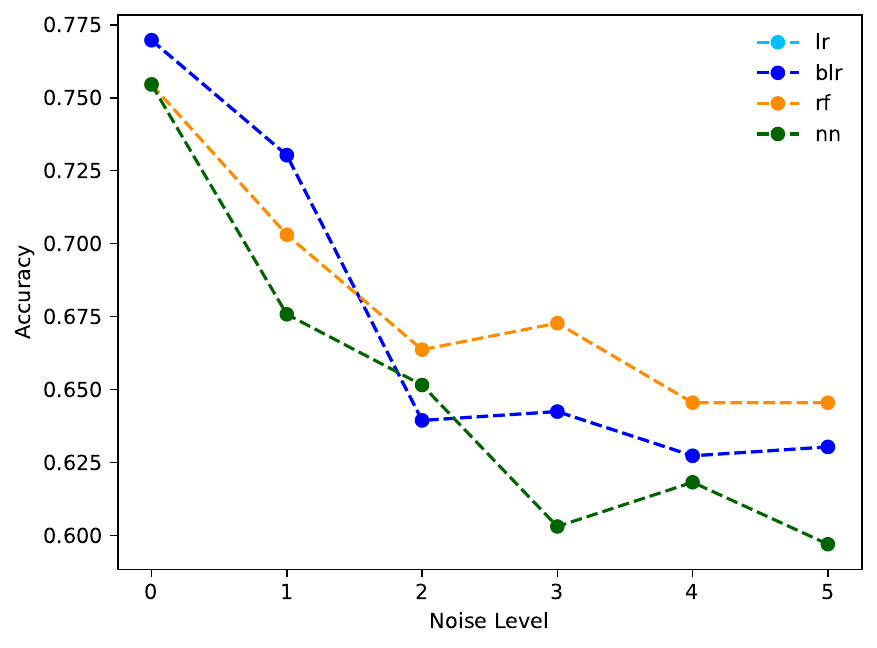}
\caption{ML models accuracy for the German Credit dataset. The \texttt{sklearn} and bayesian linear regression results are very similar and overlap in the graph. 
}
\label{fig:german_credit_accuracy}
\end{figure}

The German Credit dataset \cite{Dua:2017a} is a widely used benchmark in machine learning and credit risk modeling. It contains 1,000 loan applicants with 20 features describing socio-economic and financial attributes. The dataset is often used for binary classification, where the objective is to predict creditworthiness. The features include categorical and numerical variables such as credit amount, duration, employment status, savings, housing status, and purpose of the loan. The dataset poses challenges related to class imbalance, feature correlations, and interpretability, making it useful for testing classification models, fairness in AI, and explainability techniques.  


We followed the same process for training the ML models and estimating CE as in Section \ref{sec:aleatoric}, apart from limiting the noise levels to 5. Figure~\ref{fig:german_credit_accuracy} shows the model accuracy for all our classifiers for the different levels of ML uncertainty. Our results for this dataset are presented in Tables~\ref{table:descr_gc} and ~\ref{table:post_gc}. Intriguingly we find that for the German Credit dataset, a rather small and challenging dataset, the linear and RL CE methods are very unstable contrary to DiCE and NICE results, whose statistics of the $\ell_1$ difference distributions have similar robustness throughout.  

\subsubsection{Adult Income}



\begin{figure}[h]
\centering
\includegraphics[width=0.8\columnwidth]{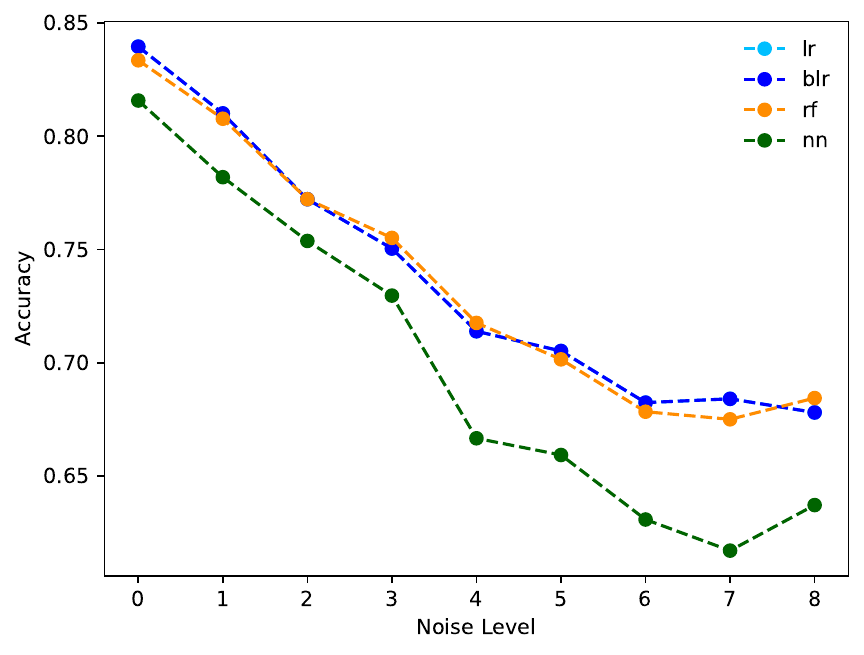}
\caption{Decrease of model accuracy for the adult income dataset as noise is increased. Bayesian logistic regression, logistic regression and random forests have very similar accuracy whereas neural networks are slightly less accurate, although we note that the architecture of the network wasn't optimized. 
}
\label{fig:ai_accuracy}
\end{figure}
The next dataset we considered is the the Adult Income \cite{Dua:2017b}, derived from the 1994 U.S. Census Bureau data, one of the most popular benchmarks for binary classification tasks. It has 14 demographic and employment-related attributes, such as age, education, occupation, hours worked per week, and marital status. The goal is to predict whether an individual’s income exceeds the threshold of \$50K per year. Due to the large number of categorical features, which considerably slow down MCMC sampling, we used a random $20\%$ subsample of the data, resulting in approximately 10,000 datapoints. During preprocessing, we also enforced a minimum frequency threshold of 300 for categorical values, filtering out infrequent categories. This preprocessing procedure was applied consistently across all our numerical experiments.

\begin{figure}[h]
\centering
\includegraphics[width=\columnwidth]{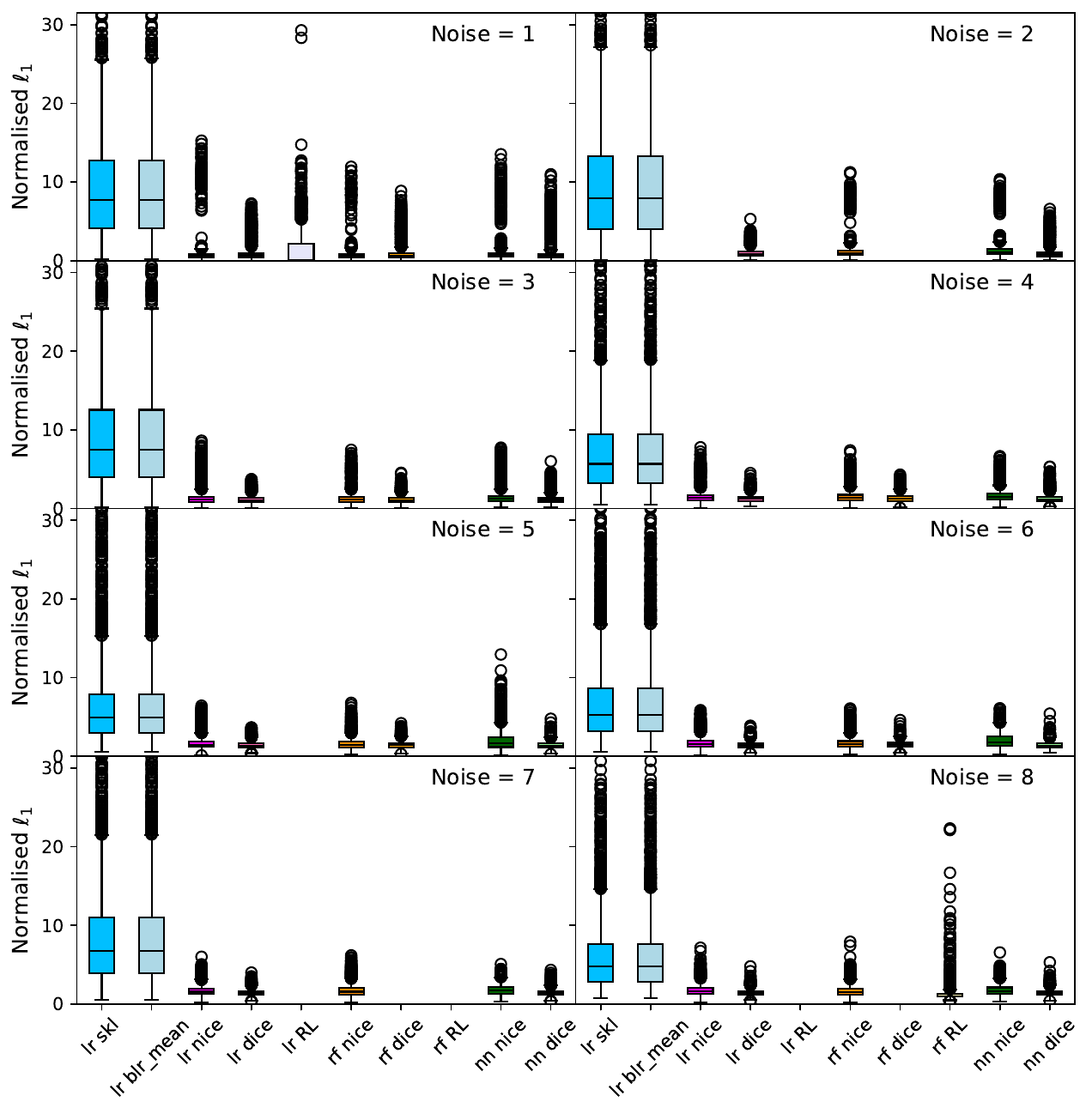}
\caption{Same as Fig. \ref{fig:l1_1_page} but for the Adult Income dateset. Note that the median and the scatter in the distributions for the linear models and models is much bigger than the corresponding for DiCE and NICE. RL CE method performed well in one simulation but not consistently. 
}
\label{fig:ai_standard}
\end{figure}

\begin{figure}[t]
\centering
\includegraphics[width=\columnwidth]{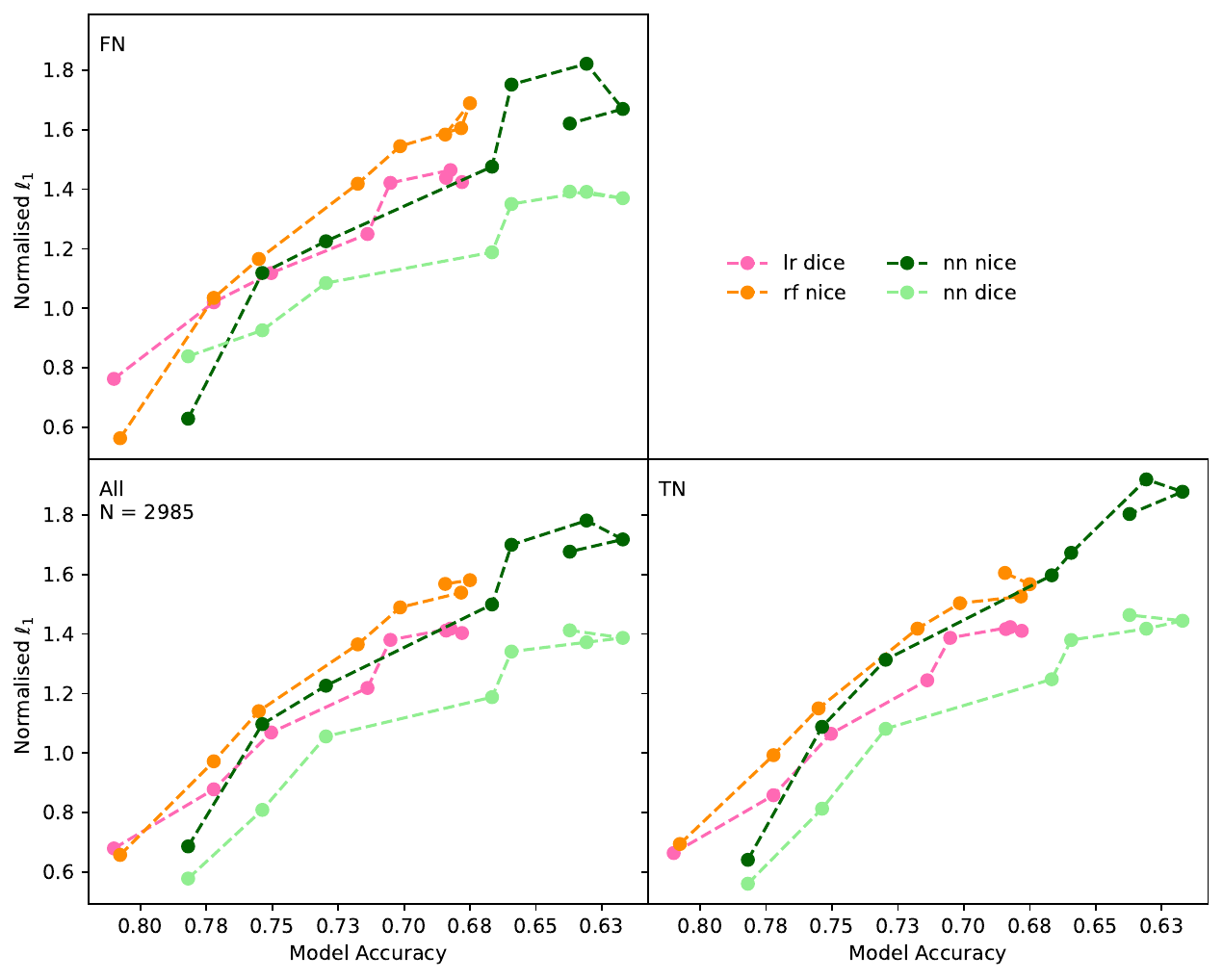}
\caption{Same as Fig. \ref{fig:accuracy_combined} but for the adult income dataset. Due to low completeness some methods are not appearing in the plots. Moreover standard and bayesian CE from linear regression have much larger medians and for clarity are not shown in the plots even though their CE are complete.}
\label{fig:adult_income_median}
\end{figure}

Fig. \ref{fig:ai_accuracy} shows the accuracy of the 4 ML algorithms we used. Even though the NN slightly underperformed in our implementation this is not a serious drawback as we are interesting in the CE. Fig. \ref{fig:ai_standard} shows the distribution of $\ell_1$ difference between the CE with minimal noise level against increasing noise. Somewhat surprisingly the median of the $\ell_1$ difference for each CE method is robust to increased noise. We also observe that apart from the LR CE models and RL whose CE are feeble, DiCE and NICE are stable with respect to the median and dispersion of their distributions. 

The adult income dataset is a staple in CE literature. From the analysis in this subsection, we judge that DiCE and NICE are the most reliable CE methods when it comes to aleatoric uncertainty robustness. Note, however, that they still present several deficiencies:

\begin{itemize}

\item{Even within the same method and model, moderately increased noise causes large variations, leading to CE error disproportional to the drop in ML accuracy. For example, in Fig. \ref{fig:adult_income_median}, the median CE change for RF with NICE increases from 36 to 46 ($\approx22\%$ change) as accuracy (Fig. \ref{fig:ai_accuracy}) drops from 0.83 to 0.80 ($\approx4\%$ decrease).}

\item {Despite comparable model accuracies between LR and RF (Fig. \ref{fig:ai_accuracy}), the DiCE-generated CE yield a ~10\% difference in median $\ell_1$ distance, as illustrated in Fig. \ref{fig:adult_income_median}}.

\item{CEs generated to explain NN models often exhibit substantial variability, even under low levels of noise injection, and this variability does not necessarily correlate with low predictive accuracy.}

\item {Some methods, despite being relatively superior, produce CE that do not consistently meet the $10\%$ completeness threshold as noise increases (e.g., LR with NICE and RF with DiCE).}

\end{itemize}

\begin{table*}[t]
\centering
\scriptsize
\setlength{\tabcolsep}{0pt}     
\begin{tabular}{llllll}
\toprule
 & Original & Noise 0 CE & Noise 1 CE & Noise 2 CE & Noise 3 CE \\
\midrule
\textbf{workclass}       & Local-gov           & State-gov         & Private          & Private          & Private        \\
\textbf{education}       & HS-grad             & HS-grad           & Bachelors        & HS-grad          & HS-grad           \\
\textbf{marital-status}  & Married-civ-spouse  & Never-married     & Divorced         & Never-married    & Married-civ-spouse \\
\textbf{occupation}      & Protective-serv     & Protective-serv   & Prof-specialty   & Adm-clerical     & Protective-serv   \\
\textbf{relationship}    & Husband             & Husband           & Husband          & Husband          & Own-child           \\
\textbf{native-country}  & usa                 & usa               & usa              & usa              & usa                   \\
\textbf{age}             & 28                  & 25                & 25               & 61               & 19                         \\
\textbf{education-num}   & 9                   & 8                 & 8                & 7                & 10                                 \\
\textbf{capital-gain}    & 0                   & 40640             & 61569            & 82140            & 81960         \\
\textbf{capital-loss}    & 0                   & 137               & 2                & 386              & 1406                 \\
\textbf{hours-per-week}  & 40                  & 44                & 39               & 51               & 47              \\
\bottomrule
\end{tabular}
\caption{The CE outputs for one FN instance of the Adult income dataset. We present the comparison between the original data point and the CE from the LR-DiCE combination with 4 levels of increasing noise. Notably features like “capital-gain”, “age” and “hours-per-week” vary significantly, whilst the model accuracy drops less than 10\%, as seen from Fig.~\ref{fig:ai_accuracy}.}
\label{tab:ai-CE-noise}
\end{table*}

It is useful to explore this volatility in more detail to better understand its impact on an individual. Table \ref{tab:ai-CE-noise} presents the CE for a FN instance from the Adult Income dataset, generated using the LR-DiCE combination. This method was selected because it demonstrated the best performance, as shown in Figures \ref{fig:ai_standard} and \ref{fig:adult_income_median}. Notably, the CE exhibit significant variation in the numerical features, while the categorical features (e.g., education and marital-status) display fluctuations that reflect instability in the counterfactual outputs.

\subsubsection{Give Me Some Credit}

\begin{figure}[H]
\centering
\includegraphics[width=1\columnwidth]{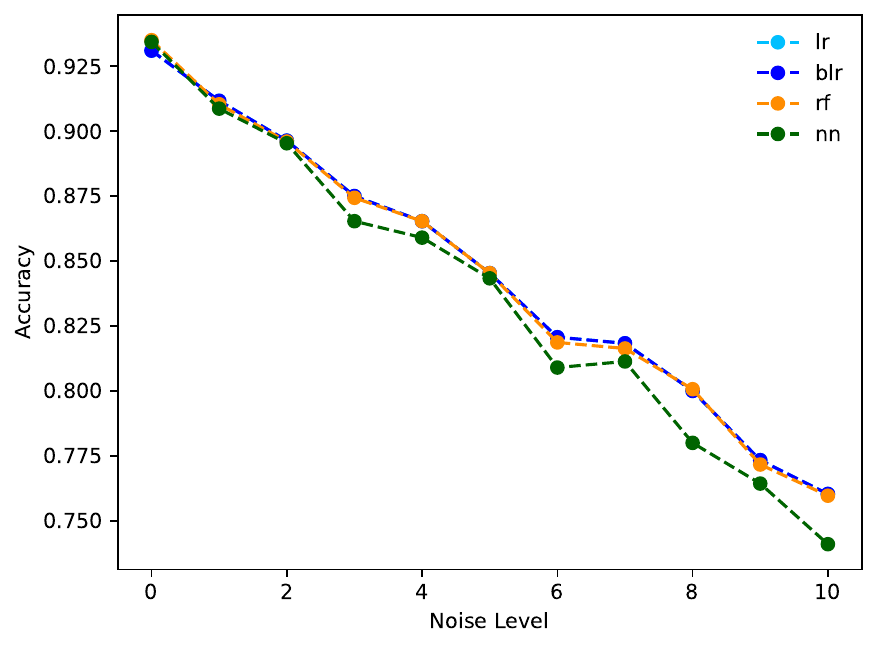}
\caption{Decrease of model accuracy for the Kaggle Give Me Some Credit dataset as noise is increased. All four ML algorithms we tried give similar results in terms of accuracy.}
\label{fig:gsc_accuracy}
\end{figure}

The “Give Me Some Credit” dataset \cite{Kaggle:2011}, originally hosted on Kaggle, is commonly used for credit risk modeling. It comprises data from approximately 120,000 credit applicants, with features capturing financial and demographic information, and aims to predict whether an individual will experience a serious delinquency (i.e., 90+ days past due) within two years. This dataset is substantially larger and more realistic than many alternatives, bearing similarities to the FICO dataset, which is another widely used benchmark in the field. It contains only numerical features, which serves as a contrast to the adult income dataset. In our analysis, we utilize only $10\%$ of the training portion of the dataset provided on Kaggle, still split into train and test set during the analysis, as it is sufficient for our purposes.

\begin{figure}[H]
\centering
\includegraphics[width=\columnwidth]{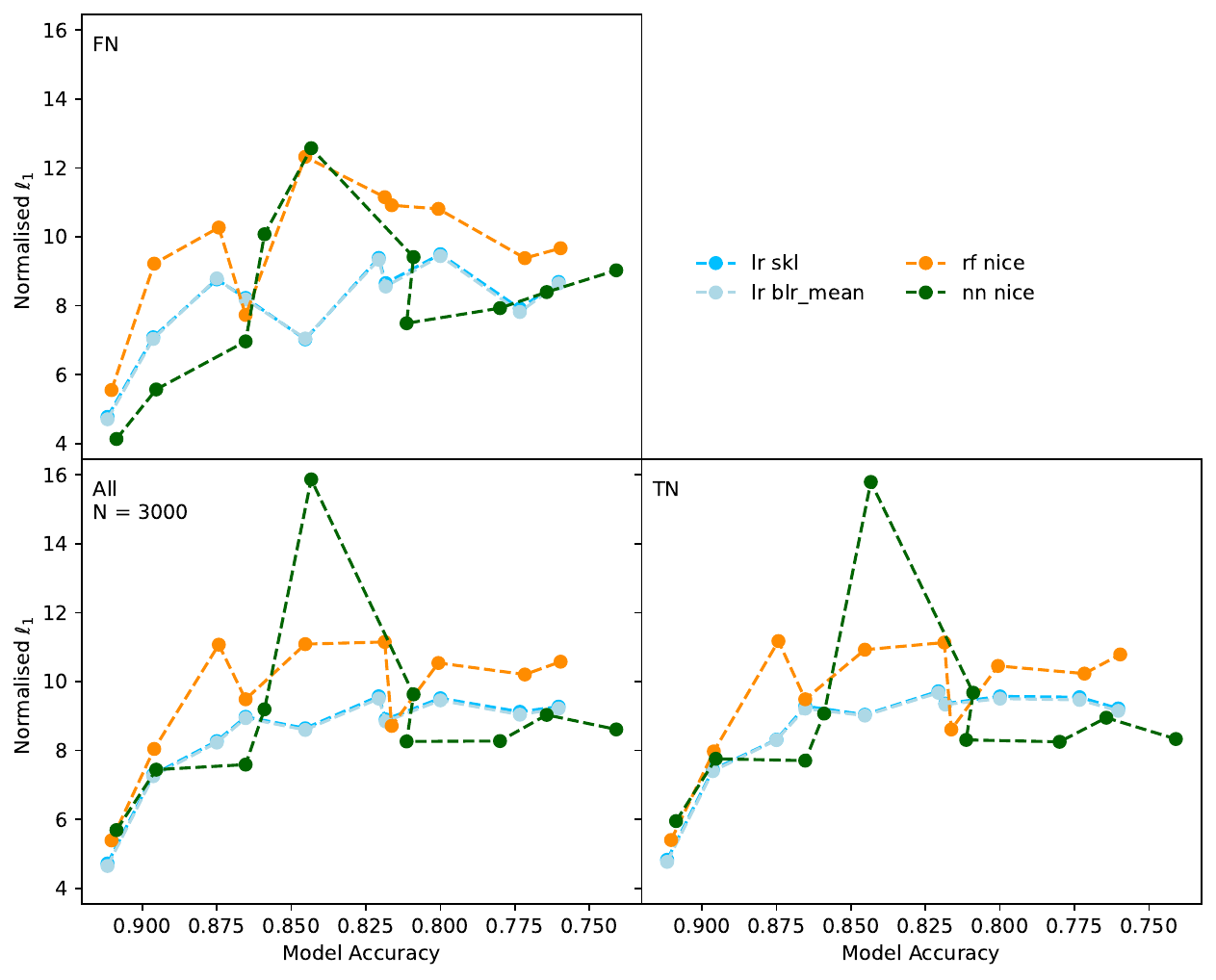}
\caption{Same as Fig. \ref{fig:accuracy_combined} but for the Give Me Some Credit dataset. Due to low completeness some methods are not appearing in the plots. Bayesian and frequentist LR have very similar results, whereas DiCE did not produced CE with sufficient completeness.   
}
\label{fig:gsc_l1}
\end{figure}

The accuracy of the 4 ML methods, as show in Fig. \ref{fig:gsc_accuracy} we considered is similar at low noise levels, but RF and LR outperforms the other methods at higher noise levels. Yet, in Fig. \ref{fig:gsc_l1} we observe that for noise levels 9 and 10 the NICE RF CE have greater median $\ell_1$ difference from the noise ground level. This raises the concern that despite the fact that the RF model exhibits better accuracy than NN and similar to LR, and therefore can be considered sufficient in this circumstance, its CE explanations are inferior when ML uncertainty increases. As a result of this dissonance it is possible that a data science practitioner, following standard practice, would choose an ML method that will, on average, be detrimental to their CE.

From Fig. \ref{fig:gsc_l1}, we observe that even methods which produce correct CE exhibit high volatility as noise increases. This suggests that the presence of a small amount of noise—typically beyond the control of data scientists—can lead to significant changes in CE, despite having only a minor impact on model accuracy. Finally, we note that although “Give Me Some Credit” is an imbalanced dataset the results the consistent for both TN and FN subsets of the test set. Still, in such settings accuracy can understate drops in minority-class performance, so we treat it as a coarse indicator of robustness.

\begin{figure}[h]
\centering
\includegraphics[width=\columnwidth]{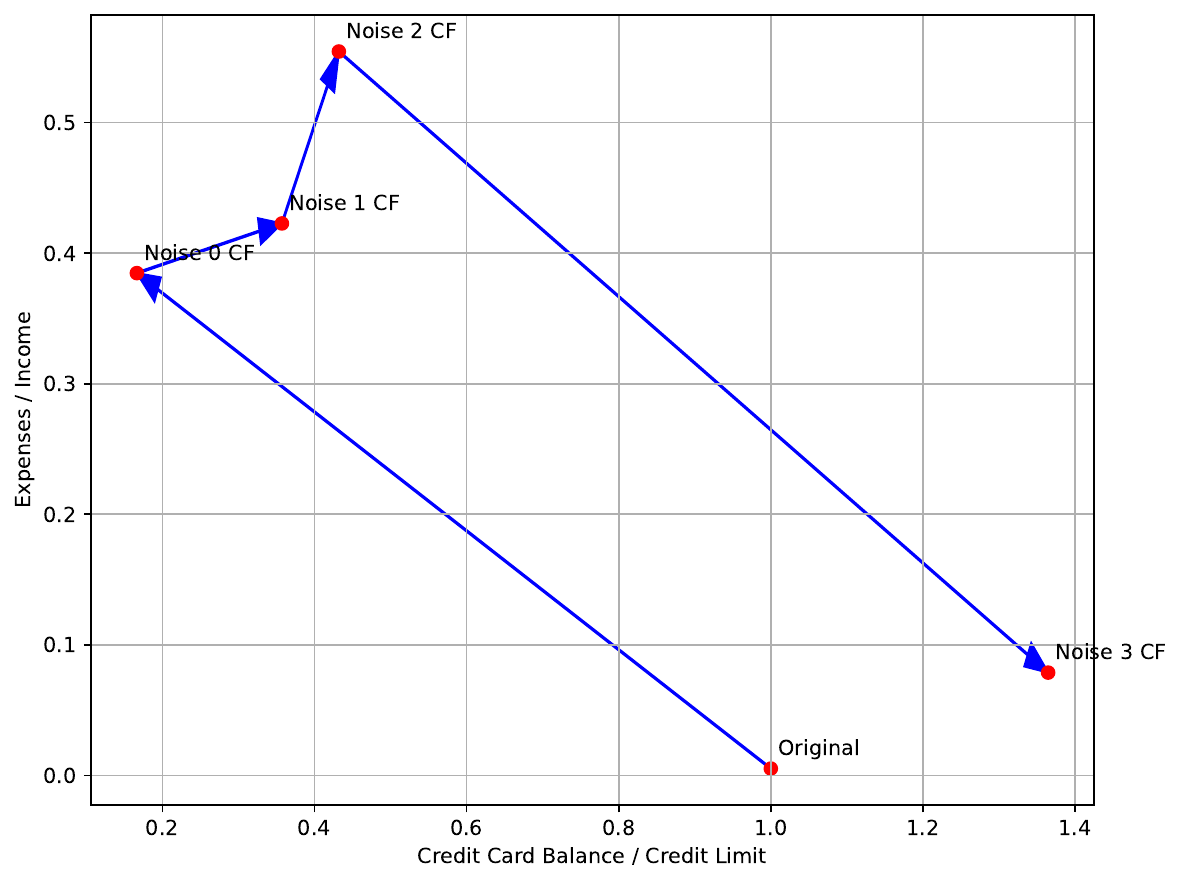}
\caption{“Give Me Some Credit” CE as input noise increases. The plot visualizes two features: (1) Total balance on credit cards and personal lines of credit (excluding real estate and installment debt) divided by the total credit limit, and (2) Monthly debt obligations (including alimony and living costs) divided by gross monthly income. These represent a 2D projection of the original 10-dimensional input space. CE were generated using the Bayesian LR model, which exhibited the lowest average variability. Interestingly, in this subspace, CE at the highest noise level appear closest to the original input.    
}
\label{fig:gsc_CE_move}
\end{figure}

Fig. \ref{fig:gsc_CE_move} shows the evolution of CE, produced using Bayesian LR, for the first (lowest) 4 noise levels. With the caveat that this is projection of a 10-dimensional space that we cannot visualize we find that in this subspace CE almost come a full circle moving further from the original instance and returning back close to it. These changes are of course modulated by other changes in the other input dimensions, however it is worth asking whether this  volatility due to $0.5$ drop in model accuracy is part of what constitutes a good explanation in ML.

\section{Conclusions}
\label{sec:conclusions}
ML classification is hugely consequential in society \cite{FOURCADE2013559}. XAI aims to provide the means of understanding its results and perhaps identifying its limitations. Our analysis herein highlights the impact of ML uncertainty on the robustness of CE when they are based solely on proximity. 

We adopted the standard distinction between aleatoric and epistemic uncertainty to systematize our analysis. Through simulations, we observed that increasing aleatoric uncertainty leads to divergent CE across different combinations of machine learning algorithms and CE methods. This variability is further influenced by dataset size and feature characteristics. Notably, CE generated by NN showed the highest robustness on the dataset with the most features, extending prior findings that deep learning often outperforms traditional ML methods in high-dimensional settings \cite{borisov2022deep}. 

Using real-world datasets and increasing their aleatoric uncertainty in a controlled fashion we uncover further disadvantages of CE, the most important of which is the disproportionate change in CE as model accuracy changes. Moreover no combination of ML algorithm and CE methods gave consistently the most robust results indicating that there are no clear choices in this space for practitioners. 

In conclusion, our results call into attention the robustness of proximity-based CE in the presence of various sources of ML uncertainty. 

\section{Acknowledgments}
LC would like to thank Constantine Dovrolis and Mihalis Nikolaou for helpful discussions and Chris Stylianou and Giorgos Kosta for help with running jobs on the HPC system Cyclone. This work was supported by computing time awarded on the Cyclone supercomputer of the High Performance Computing Facility of The Cyprus Institute under project HPCF-3350.

During the preparation of this work the authors used LLMs in order to fix grammatical errors. After using this tool, the authors reviewed and edited the content as needed and take  full responsibility for the content of the publication.


\bibliography{example_paper}
\bibliographystyle{icml2025}

\clearpage
\newpage
\appendix
\section{Statistical Results}
\label{sec:appendix}
Here we provide the results of the corresponding statistical analysis in supplementary tables and figures that were omitted from the main text for readability. 

\subsection{Synthetic Data}

\begin{table}[h]
\centering
\resizebox{\columnwidth}{!}{%
\begin{tabular}{lllllllllllll}
\toprule
 & \multicolumn{3}{r}{Median} & \multicolumn{3}{c}{P10} & \multicolumn{3}{c}{P90} & \multicolumn{3}{c}{IQR} \\
Uncertainty & High & Low & Medium & High & Low & Medium & High & Low & Medium & High & Low & Medium \\
ML-CF Method &  &  &  &  &  &  &  &  &  &  &  &  \\
\midrule
lr-RL & 0.25 & 0.04 & 0.33 & 0.16 & 0.01 & 0.21 & 0.49 & 0.11 & 0.45 & 0.18 & 0.06 & \textbf{0.11} \\
lr-blr (marg) & \textbf{0.18} & 0.00 & 0.29 & 0.09 & 0.00 & 0.10 & \textbf{0.43} & 0.00 & 0.35 & 0.16 & 0.00 & 0.12 \\
lr-blr (mean) & 0.18 & 0.00 & 0.29 & 0.09 & 0.00 & 0.10 & 0.43 & 0.00 & \textbf{0.35} & 0.16 & 0.00 & 0.12 \\
lr-dice & 1.05 & 0.82 & 0.83 & 0.31 & 0.20 & 0.23 & 2.70 & 2.16 & 2.23 & 1.11 & 0.90 & 0.97 \\
lr-nice & 0.31 & \textbf{0.00} & 0.33 & 0.12 & \textbf{0.00} & 0.13 & 0.66 & \textbf{0.00} & 0.49 & 0.28 & \textbf{0.00} & 0.19 \\
lr-skl & 0.18 & \textbf{0.00} & 0.30 & \textbf{0.09} & \textbf{0.00} & 0.11 & 0.43 & \textbf{0.00} & 0.36 & \textbf{0.16} & \textbf{0.00} & 0.12 \\
rf-RL & 0.92 & 0.05 & \textbf{0.24} & 0.50 & 0.02 & \textbf{0.07} & 1.66 & 0.13 & 0.67 & 0.48 & 0.04 & 0.37 \\
rf-dice & 1.25 & 0.82 & 0.91 & 0.41 & 0.18 & 0.27 & 2.68 & 2.05 & 2.31 & 1.16 & 0.92 & 1.04 \\
rf-nice & 0.63 & \textbf{0.00} & 0.43 & 0.21 & \textbf{0.00} & 0.13 & 1.47 & \textbf{0.00} & 0.91 & 0.57 & \textbf{0.00} & 0.41 \\
nn-dice & 1.06 & 0.94 & 1.03 & 0.34 & 0.18 & 0.30 & 2.51 & 2.31 & 2.44 & 1.09 & 1.20 & 1.14 \\
nn-nice & 0.55 & 0.20 & 0.36 & 0.30 & \textbf{0.00} & 0.10 & 1.14 & 0.36 & 0.74 & 0.31 & 0.31 & 0.37 \\
\bottomrule
\end{tabular}
}
\caption{Same as Table~\ref{table:descr1} but for mock dataset 2.}
\label{table:descr2}
\end{table}

\begin{table}[h]
\centering
\resizebox{\columnwidth}{!}{%
\begin{tabular}{lllllllllllll}
\toprule
 & \multicolumn{3}{c}{Median} & \multicolumn{3}{c}{P10} & \multicolumn{3}{c}{P90} & \multicolumn{3}{c}{IQR} \\
Uncertainty & High & Low & Medium & High & Low & Medium & High & Low & Medium & High & Low & Medium \\
ML-CF Method &  &  &  &  &  &  &  &  &  &  &  &  \\
\midrule
lr-RL & nan & nan & 1.60 & nan & nan & \textbf{0.43} & nan & nan & 8.54 & nan & nan & 2.05 \\
lr-blr (marg) & 1.17 & 0.31 & 0.98 & 0.67 & 0.07 & 0.47 & 1.79 & 0.72 & 1.61 & 0.61 & 0.38 & 0.60 \\
lr-blr (mean) & 1.17 & 0.31 & 0.98 & 0.67 & 0.07 & 0.47 & 1.79 & 0.72 & 1.61 & 0.61 & 0.38 & 0.60 \\
lr-dice & 1.31 & 0.63 & 1.12 & 0.76 & 0.28 & 0.61 & 1.92 & 1.11 & 1.74 & 0.62 & 0.42 & 0.61 \\
lr-nice & 1.15 & 0.51 & 1.03 & 0.67 & 0.12 & 0.53 & 1.76 & 1.03 & 1.61 & 0.57 & 0.50 & 0.57 \\
lr-skl & 1.17 & 0.30 & 0.98 & 0.67 & \textbf{0.07} & 0.47 & 1.78 & \textbf{0.71} & 1.60 & 0.60 & 0.38 & 0.60 \\
rf-RL & 1.76 & \textbf{0.15} & nan & \textbf{0.49} & 0.09 & nan & 2.78 & 1.28 & nan & 1.28 & \textbf{0.26} & nan \\
rf-dice & 1.19 & 0.64 & 1.09 & 0.72 & 0.28 & 0.59 & 1.80 & 1.10 & 1.64 & \textbf{0.52} & 0.44 & \textbf{0.53} \\
rf-nice & \textbf{1.09} & 0.48 & \textbf{0.95} & 0.62 & 0.15 & 0.46 & \textbf{1.68} & 1.02 & \textbf{1.54} & 0.59 & 0.50 & 0.59 \\
nn-dice & 1.20 & 0.59 & 1.08 & 0.73 & 0.24 & 0.63 & 1.79 & 1.06 & 1.68 & 0.55 & 0.40 & 0.56 \\
nn-nice & 1.23 & 0.60 & 1.08 & 0.74 & 0.21 & 0.58 & 1.81 & 1.04 & 1.70 & 0.62 & 0.42 & 0.59 \\
\bottomrule
\end{tabular}
}
\caption{Same as Table~\ref{table:descr1} but for mock dataset 3.}
\label{table:descr3}
\end{table}

\begin{table}[h]
\centering
\resizebox{\columnwidth}{!}{%
\begin{tabular}{lllllllllllll}
\toprule
 & \multicolumn{3}{c}{Median} & \multicolumn{3}{c}{P10} & \multicolumn{3}{c}{P90} & \multicolumn{3}{r}{IQR} \\
Uncertainty & High & Low & Medium & High & Low & Medium & High & Low & Medium & High & Low & Medium \\
ML-CF Method &  &  &  &  &  &  &  &  &  &  &  &  \\
\midrule
lr-blr (mean) & \textbf{1.19} & \textbf{0.22} & 0.89 & \textbf{0.95} & 0.10 & \textbf{0.66} & 1.47 & \textbf{0.41} & 1.18 & 0.26 & \textbf{0.16} & 0.27 \\
lr-dice & 1.24 & 0.40 & 1.01 & 0.99 & 0.22 & 0.76 & 1.53 & 0.61 & 1.29 & 0.27 & 0.21 & 0.27 \\
lr-nice & 1.23 & 0.33 & 0.98 & 0.97 & 0.17 & 0.73 & 1.52 & 0.55 & 1.28 & 0.27 & 0.19 & 0.28 \\
lr-skl & 1.19 & 0.22 & \textbf{0.89} & 0.95 & \textbf{0.10} & 0.66 & \textbf{1.47} & 0.41 & \textbf{1.18} & 0.26 & 0.16 & 0.27 \\
rf-nice & 1.23 & 0.39 & 1.01 & 0.98 & 0.20 & 0.75 & 1.54 & 0.73 & 1.31 & 0.28 & 0.28 & 0.29 \\
nn-dice & 1.23 & 0.41 & 1.00 & 0.99 & 0.26 & 0.76 & 1.50 & 0.63 & 1.27 & \textbf{0.26} & 0.20 & \textbf{0.27} \\
nn-nice & 1.22 & 0.38 & 0.99 & 0.98 & 0.22 & 0.74 & 1.51 & 0.62 & 1.29 & 0.28 & 0.20 & 0.28 \\
\bottomrule
\end{tabular}
}
\caption{Same as Table~\ref{table:descr1} but for mock dataset 4.}
\label{table:descr4}
\end{table}

\begin{table}[h]
\centering
\resizebox{\columnwidth}{!}{%
\begin{tabular}{lllllll}
\toprule
 & \multicolumn{2}{c}{Median Delta} & \multicolumn{2}{c}{Posterior P(best)} & \multicolumn{2}{c}{Significance} \\
Noise & High & Medium & High & Medium & High & Medium \\
Method &  &  &  &  &  &  \\
\midrule
lr-RL & NaN & -0.14 & NaN & 1.0 & NaN &  \\
lr-blr (marg) & -0.01 & NaN & 0.73 & NaN &  &  \\
lr-blr (mean) & -0.01 & -0.00 & 0.73 & 0.51 &  &  \\
lr-dice & -0.21 & -0.55 & 1.0 & 1.0 &  & *** \\
lr-nice & -0.22 & -0.45 & 1.0 & 1.0 &  & *** \\
lr-skl & -0.01 & NaN & 0.72 & NaN &  & NaN \\
rf-RL & -0.05 & -0.17 & 1.0 & 1.0 &  &  \\
rf-dice & -0.22 & -0.51 & 1.0 & 1.0 &  &  \\
rf-nice & -0.06 & -0.35 & 1.0 & 1.0 &  & *** \\
tf-dice & -0.26 & -0.50 & 1.0 & 1.0 &  & *** \\
tf-nice & -0.23 & -0.41 & 1.0 & 1.0 &  & *** \\
\bottomrule
\end{tabular}
}
\caption{Same as Table~\ref{table:p1} but for mock dataset 2}
\label{table:p2}
\end{table}

\begin{table}[h]
\centering
\resizebox{\columnwidth}{!}{%
\begin{tabular}{llllllllll}
\toprule
 & \multicolumn{3}{r}{Median Delta} & \multicolumn{3}{r}{Posterior P(best)} & \multicolumn{3}{r}{Significance} \\
Noise & High & Low & Medium & High & Low & Medium & High & Low & Medium \\
Method &  &  &  &  &  &  &  &  &  \\
\midrule
lr-RL & NaN & NaN & -0.72 & NaN & NaN & 1.0 &  &  &  \\
lr-blr (marg) & -0.08 & -0.12 & -0.03 & 1.0 & 1.0 & 0.92 & *** &  & *** \\
lr-blr (mean) & -0.08 & -0.12 & -0.03 & 1.0 & 1.0 & 0.92 & *** &  & *** \\
lr-dice & -0.22 & -0.45 & -0.17 & 1.0 & 1.0 & 1.0 & *** &  & *** \\
lr-nice & -0.06 & -0.32 & -0.06 & 0.99 & 1.0 & 0.99 & *** &  & *** \\
lr-skl & -0.08 & -0.11 & -0.02 & 1.0 & 1.0 & 0.87 & *** &  & ** \\
rf-RL & -0.68 & NaN & NaN & 1.0 & NaN & NaN &  & NaN &  \\
rf-dice & -0.11 & -0.47 & -0.13 & 1.0 & 1.0 & 1.0 &  &  &  \\
rf-nice & NaN & -0.30 & NaN & NaN & 1.0 & NaN & NaN &  & NaN \\
tf-dice & -0.12 & -0.41 & -0.13 & 1.0 & 1.0 & 1.0 & *** &  & *** \\
tf-nice & -0.14 & -0.41 & -0.12 & 1.0 & 1.0 & 1.0 & *** &  & *** \\
\bottomrule
\end{tabular}
}
\caption{Same as Table~\ref{table:p1} but for mock dataset 3}
\label{table:p3}
\end{table}

\begin{table}[h]
\centering
\resizebox{\columnwidth}{!}{%
\begin{tabular}{llllllllll}
\toprule
 & \multicolumn{3}{r}{Median Delta} & \multicolumn{3}{r}{Posterior P(best)} & \multicolumn{3}{r}{Significance} \\
Noise & High & Low & Medium & High & Low & Medium & High & Low & Medium \\
Method &  &  &  &  &  &  &  &  &  \\
\midrule
lr-blr (mean) & NaN & NaN & 0.00 & NaN & NaN & 0.49 & NaN & NaN &  \\
lr-dice & -0.05 & -0.17 & -0.11 & 1.0 & 1.0 & 1.0 & *** & *** & *** \\
lr-nice & -0.03 & -0.11 & -0.09 & 1.0 & 1.0 & 1.0 & *** & *** & *** \\
lr-skl & 0.00 & -0.00 & NaN & 0.49 & 0.50 & NaN & ** & *** & NaN \\
rf-nice & -0.04 & -0.18 & -0.12 & 1.0 & 1.0 & 1.0 & *** & *** & *** \\
tf-dice & -0.04 & -0.19 & -0.10 & 1.0 & 1.0 & 1.0 & *** & *** & *** \\
tf-nice & -0.04 & -0.16 & -0.10 & 1.0 & 1.0 & 1.0 & *** & *** & *** \\
\bottomrule
\end{tabular}
}
\caption{Same as Table~\ref{table:p1} but for mock dataset 4}
\label{table:p4}
\end{table}

\subsection{Real World Data}

\begin{table}[h]
\centering
\resizebox{\columnwidth}{!}{%
\begin{tabular}{lllllllll}
\toprule
 & \multicolumn{2}{c}{Median} & \multicolumn{2}{c}{P10} & \multicolumn{2}{c}{P90} & \multicolumn{2}{c}{IQR} \\
Uncertainty & Low & Medium & Low & Medium & Low & Medium & Low & Medium \\
ML-CF Method &  &  &  &  &  &  &  & \\
\midrule
lr-RL & 0.87 & 1.36 & 0.36 & 1.03 & 3.04 & 1.85 & 0.91 & 0.39 \\
lr-blr (mean) & 20.99 & 7.10 & 4.37 & 2.13 & 94.90 & 16.04 & 28.38 & 7.40 \\
lr-dice & 0.57 & 1.13 & 0.34 & 0.87 & 0.87 & 1.49 & 0.26 & 0.32 \\
lr-nice & \textbf{0.50} & 1.01 & 0.28 & 0.73 & \textbf{0.77} & 1.29 & 0.26 & \textbf{0.30} \\
lr-skl & 20.98 & 7.10 & 4.37 & 2.13 & 94.88 & 16.02 & 28.35 & 7.41 \\
rf-RL & 1.35 & 1.31 & 0.78 & 1.00 & 2.98 & 7.44 & 0.62 & 0.93 \\
rf-dice & 0.56 & 1.12 & 0.34 & 0.80 & 0.91 & 1.45 & \textbf{0.25} & 0.33 \\
rf-nice & 0.51 & \textbf{0.93} & \textbf{0.28} & \textbf{0.70} & 0.86 & \textbf{1.23} & 0.31 & 0.30 \\
tf-dice & 0.52 & 0.97 & 0.33 & 0.74 & 0.83 & 1.32 & 0.26 & 0.31 \\
tf-nice & 0.56 & 1.04 & 0.31 & 0.77 & 0.90 & 1.40 & 0.31 & 0.32 \\
\bottomrule
\end{tabular}
}
\caption{German credit}
\label{table:descr_gc}
\end{table}

\begin{table}[ht]
\centering
\resizebox{\columnwidth}{!}{%
\begin{tabular}{lllllll}
\toprule
 & \multicolumn{2}{c}{Median Delta} & \multicolumn{2}{c}{Posterior P(best)} & \multicolumn{2}{c}{Significance} \\
Noise & Low & Medium & Low & Medium & Low & Medium \\
Method  &  &  &  &  &  &  \\
\midrule
lr-RL & -0.33 & -0.42 & 1.0 & 1.0  &  &  \\
lr-blr (mean) & -19.42 & -6.13 & 1.0 & 1.0  &  & *** \\
lr-dice & -0.09 & -0.20 & 1.0 & 1.0  & *** & *** \\
lr-nice & NaN & -0.06 & NaN & 0.99 & NaN & *** \\
lr-skl & -19.42 & -6.13 & 1.0 & 1.0 &  & *** \\
rf-RL & -0.83 & -0.40 & 1.0 & 1.0 & *** & *** \\
rf-dice & -0.07 & -0.18 & 1.0 & 1.0 &  &  \\
rf-nice & -0.03 & NaN & 0.98 & NaN & ** & NaN \\
tf-dice & -0.04 & -0.05 & 0.995 & 0.99 & *** & *** \\
tf-nice & NaN & -0.06 & -0.09 & 1.0 & 1.0 & *** \\
\bottomrule
\end{tabular}
}
\caption{Same as Table~\ref{table:p1} but for the German Credit dataset}
\label{table:post_gc}
\end{table}

\begin{table}[ht]
\centering
\resizebox{\columnwidth}{!}{%
\begin{tabular}{lllllllllllll}
\toprule
 & \multicolumn{3}{c}{Median} & \multicolumn{3}{c}{P10} & \multicolumn{3}{c}{P90} & \multicolumn{3}{c}{IQR} \\
Uncertainty & High & Low & Medium & High & Low & Medium & High & Low & Medium & High & Low & Medium \\
ML-CF Method &  &  &  &  &  &  &  &  &  &  &  &  \\
\midrule
lr-RL & nan & \textbf{0.09} & nan & nan & \textbf{0.03} & nan & nan & 6.36 & nan & nan & 2.10 & nan \\
lr-blr (mean) & 4.80 & 7.68 & 5.67 & 1.91 & 1.96 & 1.90 & 10.99 & 19.17 & 14.27 & 4.71 & 8.63 & 6.26 \\
lr-dice & 1.40 & 0.68 & 1.22 & 0.97 & 0.22 & \textbf{0.74} & 1.96 & 1.49 & 1.88 & 0.49 & 0.55 & 0.56 \\
lr-nice & 1.62 & 0.59 & 1.32 & 1.03 & 0.20 & 0.76 & 2.72 & 1.10 & 2.16 & 0.81 & 0.47 & 0.66 \\
lr-skl & 4.80 & 7.68 & 5.67 & 1.91 & 1.96 & 1.90 & 10.99 & 19.16 & 14.27 & 4.71 & 8.63 & 6.26 \\
rf-RL & \textbf{1.29} & nan & nan & \textbf{0.85} & nan & nan & \textbf{1.45} & nan & nan & \textbf{0.35} & nan & nan \\
rf-dice & nan & 0.67 & 1.23 & nan & 0.24 & 0.76 & nan & 1.48 & 1.99 & nan & 0.52 & 0.58 \\
rf-nice & 1.57 & 0.66 & 1.36 & 0.98 & 0.25 & 0.81 & 2.53 & 1.14 & 2.21 & 0.79 & 0.47 & 0.69 \\
tf-dice & 1.41 & 0.58 & \textbf{1.19} & 0.98 & 0.20 & 0.75 & 1.96 & \textbf{1.02} & \textbf{1.82} & 0.48 & \textbf{0.42} & \textbf{0.51} \\
nn-nice & 1.68 & 0.69 & 1.50 & 1.07 & 0.32 & 0.89 & 2.60 & 1.26 & 2.36 & 0.79 & 0.46 & 0.73 \\
\bottomrule
\end{tabular}
}
\caption{Same as Table~\ref{table:descr1} but for the Adult Income dataset.}
\label{table:ai_desc}
\end{table}

\begin{table}[ht]
\centering
\resizebox{\columnwidth}{!}{%
\begin{tabular}{llllllllll}
\toprule
 & \multicolumn{3}{c}{Median Delta} & \multicolumn{3}{c}{Posterior P(best)} & \multicolumn{3}{c}{Significance} \\
Noise & High & Low & Medium & High & Low & Medium & High & Low & Medium \\
Method &  &  &  &  &  &  &  &  &  \\
\midrule
lr-RL & NaN & NaN & NaN & NaN & NaN & NaN &  & NaN &  \\
lr-blr (mean) & -3.70 & -7.44 & -4.59 & 1.0 & 1.0 & 1.0 &  &  & *** \\
lr-dice & -0.20 & -0.39 & -0.03 & 1.0 & 1.0 & 0.99 &  &  &  \\
lr-nice & -0.44 & -0.32 & -0.12 & 1.0 & 1.0 & 1.0 &  &  &  \\
lr-skl & -3.70 & -7.44 & -4.59 & 1.0 & 1.0 & 1.0 &  &  & *** \\
rf-RL & NaN & NaN & NaN & NaN & NaN & NaN & NaN &  &  \\
rf-dice & NaN & -0.38 & -0.05 & NaN & 1.0 & 1.0 &  &  &  \\
rf-nice & -0.38 & -0.38 & -0.18 & 1.0 & 1.0 & 1.0 &  &  &  \\
nn-dice & -0.21 & -0.30 & NaN & 1.0 & 1.0 & NaN &  &  & NaN \\
nn-nice & -0.49 & -0.41 & -0.31 & 1.0 & 1.0 & 1.0 &  &  & *** \\
\bottomrule
\end{tabular}
}
\caption{Same as Table~\ref{table:p1} but for the Adult Income dataset.}
\label{table:post_ai}
\end{table}

\begin{table}[ht]
\centering
\resizebox{\columnwidth}{!}{%
\begin{tabular}{lllllllllllll}
\toprule
 & \multicolumn{3}{c}{Median} & \multicolumn{3}{c}{P10} & \multicolumn{3}{c}{P90} & \multicolumn{3}{c}{IQR} \\
Uncertainty & High & Low & Medium & High & Low & Medium & High & Low & Medium & High & Low & Medium \\
ML-CF Method &  &  &  &  &  &  &  &  &  &  &  &  \\
\midrule
lr-blr (mean) & 9.46 & 4.65 & \textbf{8.94} & \textbf{3.85} & 1.99 & 3.64 & 18.71 & 8.73 & 17.23 & 7.95 & 3.56 & 7.55 \\
lr-nice & nan & 6.33 & nan & nan & 3.01 & nan & nan & 11.09 & nan & nan & 4.27 & nan \\
lr-skl & 9.52 & 4.72 & 8.98 & 3.87 & 2.02 & 3.68 & 18.72 & 8.82 & 17.27 & 8.01 & 3.61 & 7.50 \\
rf-RL & nan & \textbf{2.15} & nan & nan & \textbf{1.27} & nan & nan & \textbf{3.48} & nan & nan & \textbf{1.15} & nan \\
rf-nice & 10.54 & 5.39 & 9.49 & 4.37 & 2.30 & 4.58 & 19.66 & 10.13 & \textbf{16.95} & 8.02 & 4.15 & 6.81 \\
tf-nice & \textbf{8.28} & 5.69 & 9.20 & 4.12 & 2.36 & \textbf{2.97} & \textbf{17.81} & 8.28 & 17.74 & \textbf{6.93} & 3.27 & \textbf{6.63} \\
\bottomrule
\end{tabular}
}
\caption{Same as Table~\ref{table:descr1} but for the Give Me Some Credit.}
\label{table:post_ai}
\end{table}

\begin{table}[ht]
\centering
\resizebox{\columnwidth}{!}{%
\begin{tabular}{llllllllll}
\toprule
 & \multicolumn{3}{c}{Median Delta} & \multicolumn{3}{c}{Posterior P(best)} & \multicolumn{3}{c}{Significance} \\
Noise & High & Low & Medium & High & Low & Medium & High & Low & Medium \\
Method &  &  &  &  &  &  &  &  &  \\
\midrule
lr-blr (mean) & -0.89 & -2.53 & NaN & 1.0 & 1.0 & NaN & *** &  & NaN \\
lr-nice & NaN & -4.24 & NaN & NaN & 1.0 & NaN &  &  &  \\
lr-skl & -0.95 & -2.59 & -0.05 & 1.0 & 1.0 & 0.65 & *** &  & *** \\
rf-RL & NaN & NaN & NaN & NaN & NaN & NaN &  & NaN &  \\
rf-nice & -1.96 & -3.29 & -0.55 & 1.0 & 1.0 & 1.0 & *** &  & *** \\
nn-nice & NaN & -3.41 & -0.17 & NaN & 1.0 & 0.89 & NaN &  & * \\
\bottomrule
\end{tabular}
}
\caption{Same as Table~\ref{table:p1} but for the Give Me Some Credit dataset.}
\label{table:post_ai}
\end{table}


\end{document}